\documentclass{article}
\usepackage{arxiv}
\usepackage[numbers]{natbib}
\usepackage[utf8]{inputenc} 
\usepackage[colorlinks=true,linkcolor=red,citecolor=blue,urlcolor=black]{hyperref}       
\usepackage{hyperref} 
\usepackage{url}            
\usepackage{booktabs}       
\usepackage{amsfonts}       
\usepackage{nicefrac}       
\usepackage{microtype}      
\usepackage[table, dvipsnames]{xcolor}         
\usepackage{caption}
\usepackage{siunitx}
\usepackage{multirow}
\usepackage{tabulary}
\usepackage{longtable}
\usepackage{graphicx}
\usepackage{subcaption}
\usepackage[most]{tcolorbox}
\tcbuselibrary{skins,breakable}
\usetikzlibrary{shadings,shadows}
\usepackage[linesnumbered]{algorithm2e}
\SetKwComment{Comment}{$\triangleright$\ }{}
\usepackage[export]{adjustbox}
\usepackage{mathtools}
\usepackage{xcolor}
\usepackage{pifont}
\usepackage{wrapfig}
\usepackage{tabularx,ragged2e}
\newcolumntype{C}{>{\Centering\arraybackslash}X} 
\usepackage{tabularx}
\usepackage[german,greek,russian, english]{babel}
\newcolumntype{P}[1]{>{\centering\arraybackslash}p{#1}}
\newtheorem{proposition}{Proposition}[section]

\newcommand{\ie}{\textit{i}.\textit{e}. }
\newcommand{\ieending}{\textit{i}.\textit{e}.}
\newcommand{\dd}[2]{\frac{\partial #1}{\partial #2}}

\newcommand{\R}{\mathcal{R}}
\newcommand{\Xbrace}[2]{{\color{orange} \overbrace{\color{black} #1}^{#2} }}
\newcommand{\xbrace}[2]{{\color{orange} \underbrace{\color{black} #1}_{#2} }}
\newcommand{\ybrace}[2]{{\color{red} \underbrace{\color{black} #1}_{#2} }}
\newcommand{\ebrace}[2]{{\color{gray} \underbrace{\color{black} #1}_{#2} }}
\newcommand{\cmark}{\textcolor{ForestGreen!80!black}{\ding{51}}}
\newcommand{\xmark}{\textcolor{red}{\ding{55}}}
\newcommand{\block}[1]{%
  \raisebox{\dimexpr(\fontcharht\font`X-1em)/2}{\rule{0.5em}{#1\dimexpr0.5em/8}}%
}
\usepackage{tcolorbox}
\newcommand{\code}[1]{\mbox{
    \ttfamily
    \tcbox[
        on line,
        boxsep=0pt, left=4pt, right=4pt, top=2pt, bottom=1.5pt,
        toprule=0pt, rightrule=0pt, bottomrule=0pt, leftrule=0pt,
        oversize=0pt, enlarge left by=0pt, enlarge right by=0pt,
        colframe=white, colback=black!12
    ]{#1}
}}
\DeclareUnicodeCharacter{2581}{\block{1}}

\title{MambaLRP: Explaining Selective State Space Sequence Models}

\author{
Farnoush Rezaei Jafari$^{1,2}$\footnotemark[1]  \quad Grégoire Montavon$^{3,2,1}$  \quad Klaus-Robert Müller$^{1,2,4,5,6}$  \quad Oliver Eberle$^{1,2}$\footnotemark[1]\\
\\
\footnotesize{$^{1}$Machine Learning Group, Technische Universit\"at Berlin, 10587 Berlin, Germany}\\
\footnotesize{$^{2}$BIFOLD -- Berlin Institute for the Foundations of Learning and Data, 10587 Berlin, Germany}\\
\footnotesize{$^{3}$Department of Mathematics and Computer Science, Freie Universit\"at Berlin,}\\
\footnotesize{Arnimallee 14, 14195 Berlin, Germany}\\
\footnotesize{$^{4}$Department of Artificial Intelligence, Korea University, Seoul 136-713, South Korea}\\
\footnotesize{$^{5}$Max Planck Institute for Informatics, Stuhlsatzenhausweg 4, 66123 Saarbrücken, Germany}\\
\footnotesize{$^{6}$Google DeepMind, Berlin, Germany}\\
\footnotesize{$^{*}$Correspondence to: rezaeijafari@campus.tu-berlin.de, oliver.eberle@tu-berlin.de}
}

\begin{document}
\maketitle
\thispagestyle{empty}
\newcommand\noticestring{%
  \footnotesize 38th Conference on Neural Information Processing Systems (NeurIPS 2024).}
\newcommand\copyrightnotice{%
\begin{tikzpicture}[remember picture,overlay]
\node[anchor=south, xshift=-92, yshift=35pt] at (current page.south) {\noticestring};
\end{tikzpicture}%
}

\copyrightnotice

\begin{abstract}
Recent sequence modeling approaches using selective state space sequence models, referred to as Mamba models, have seen a surge of interest. These models allow efficient processing of long sequences in linear time and are rapidly being adopted in a wide range of applications such as language modeling, demonstrating promising performance. To foster their reliable use in real-world scenarios, it is crucial to augment their transparency. Our work bridges this critical gap by bringing explainability, particularly Layer-wise Relevance Propagation (LRP), to the Mamba architecture. Guided by the axiom of relevance conservation, we identify specific components in the Mamba architecture, which cause unfaithful explanations. To remedy this issue, we propose MambaLRP, a novel algorithm within the LRP framework, which ensures a more stable and reliable relevance propagation through these components. Our proposed method is theoretically sound and excels in achieving state-of-the-art explanation performance across a diverse range of models and datasets. Moreover, MambaLRP facilitates a deeper inspection of Mamba architectures, uncovering various biases and evaluating their significance. It also enables the analysis of previous speculations regarding the long-range capabilities of Mamba models.

\end{abstract}
\section{Introduction}
\label{sec:introduction}
Sequence modeling has demonstrated its effectiveness and versatility across a wide variety of tasks and data types, including text, time series, genomics, audio, and computer vision \cite{devlin-etal-2019-bert, haoyietal-informer-2021, gong21b_interspeech, avsec2021effective, dosovitskiy2021an}. Recently, there has been a surge of interest in a new class of sequence modeling architectures, known as structured state space sequence models (SSMs) \citep{gu2022efficiently, smith2023simplified, mamba}. This is due to their ability to process sequences in linear time, as opposed to quadratic time required by the more established Transformer architectures \cite{transformer}. The recent Mamba architecture, a prominent and widely adopted instance of state space models, has demonstrated competitive predictive performance on a variety of sequence modeling tasks across domains and applications \citep{mamba, Ma2024UMambaEL, mambaViT2, mambaVideo, mambaGraph1}, while scaling linearly with sequence length.

As Mamba models, and more generally SSMs, are rapidly being adopted into real-world applications, ensuring their transparency is crucial. This enables inspection beyond test set accuracy and uncovering various forms of biases, including `Clever-Hans' effects \citep{lapuschkin-ncomm19}. It is particularly important in high-risk domains such as medicine, where the prediction behavior must be robust under real-world conditions and aligned with human understanding. The field of Explainable AI \cite{montavon2018methods, gunning2019,arietta2020,DBLP:journals/pieee/SamekMLAM21} focuses on developing faithful model explanations that attribute predictions to relevant features and has shown success in explaining many highly nonlinear models such as convolutional networks \cite{chormai-tpami24}, or attention-based Transformer models \cite{DBLP:conf/icml/AliSEMMW22,achtibat2024attnlrp}. 
\begin{figure}[!t]
    \centering
    \includegraphics[width=\textwidth]{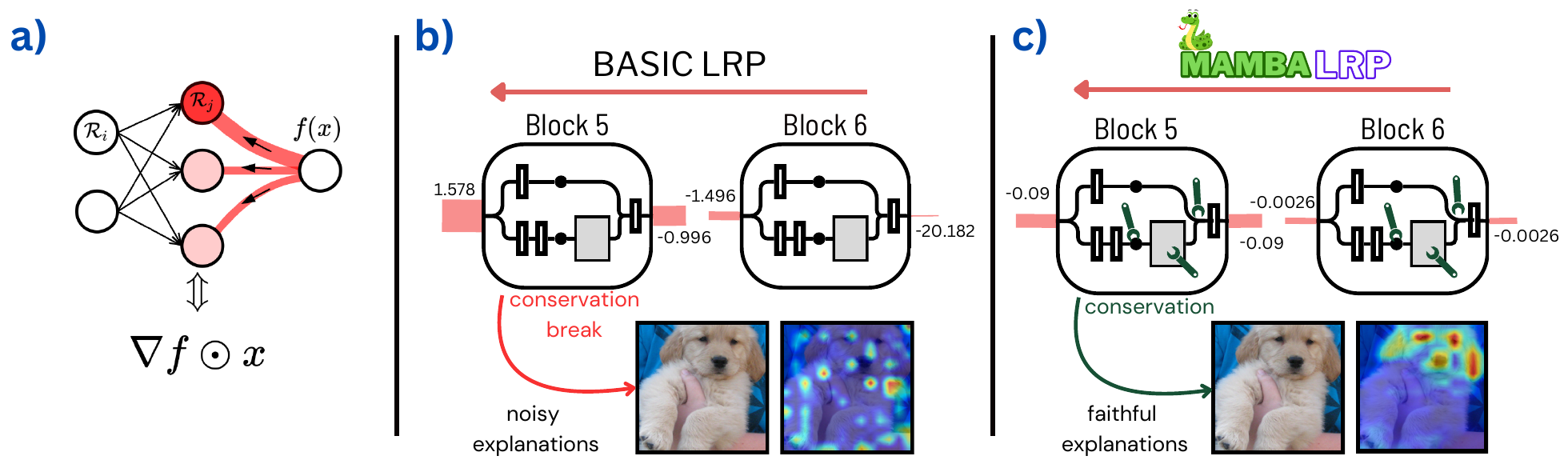}
    \caption{Conceptual steps involved in the design of MambaLRP. (a) Take as a starting point a basic LRP procedure, equivalent to Gradient$\,\times\,$Input. (b) Analyze layers in which the conservation property is violated. (c) Rework the relevance propagation strategy at those layers to achieve conservation. The resulting MambaLRP method enables efficient and faithful explanations.}
    \label{fig:overview}
    \vspace{-10pt}
\end{figure}

Explaining the predictions of Mamba models is however challenging due to their highly non-linear and recurrent structure. A recent study \cite{ali2024hidden} suggests viewing these models as attention-based models, enabling the use of attention-based explanation methods \cite{abnar-zuidema-2020-quantifying, chefer2021transformer}. Yet, the explanations produced by attention-based techniques are often unreliable and exposed to potential misalignment between input features and attention scores \cite{wiegreffe-pinter-2019-attention, jain-wallace-2019-attention}. As an alternative, Layer-wise Relevance Propagation (LRP) \cite{bach-plos15} decomposes the model function with the goal of explicitly identifying the relevance of input features by applying purposely designed propagation rules at each layer. A distinguishing feature of LRP is its adherence to a conservation axiom, which prevents the artificial amplification or suppression of feature relevance in the backward pass. LRP has been demonstrated to produce faithful explanations across various domains (e.g.\ \cite{Arras2019, schnake2022higher, DBLP:conf/icml/AliSEMMW22, chormai-tpami24}). Nevertheless, the peculiarities of the Mamba architecture are not addressed by the existing LRP procedures, which may lead to the violation of the conservation property and result in unreliable explanations.

In this work, we present MambaLRP, a novel approach to integrate LRP into the Mamba architecture. By examining the relevance propagation process across Mamba layers through the lens of conservation, we pinpoint layers within the Mamba architecture that need to be addressed specifically. We propose a novel relevance propagation strategy for these layers, grounded in the conservation axiom, that is theoretically sound, straightforward to implement and computationally efficient. Through a number of quantitative evaluations, we show that the proposed MambaLRP approach allows to robustly deliver the desired high explanatory performance, exceeding by far the performance of various baseline explanation methods as well as a naive transposition of LRP to the Mamba architecture. We further demonstrate the usefulness of MambaLRP in several areas: gaining concrete insights into the model's prediction mechanism, uncovering undesired decision strategies in image classification, identifying gender bias in language models, and analyzing the long-range capabilities of Mamba. Our code is publicly available.\footnote{\url{https://github.com/FarnoushRJ/MambaLRP}}

\section{Related Work}
\paragraph*{Structured State Space Sequence Models (SSMs).} Transformers \cite{transformer} have emerged as the most widely used architectures for sequence modeling. However, their computational limitations, particularly with large sequence lengths, have restricted their applicability in modeling long sequences. Addressing these computational limitations, recent works \citep{gu2021combining, gu2022efficiently} have introduced structured state space sequence models (SSMs) as an alternative approach. SSMs are a class of sequence modeling methods, leveraging the strengths of recurrent, convolutional, and continuous-time methods, demonstrating promising performance across various domains, including language~\cite{fu2022hungry, mehta2022long}, image~\cite{yan2023diffusion, baron20232, nguyen2022s4nd}, and video~\cite{wang2023selective} processing, and beyond~\cite{saon2023diagonal, david2022decision, lu2024structured}. A recent advancement by \citet{mamba} introduced selective SSM, an enhanced data-dependent SSM with a selection mechanism that adjusts its parameters based on the input. Built on this dynamic selection, the Mamba architecture fuses the SSM components with multilayer perceptron (MLP) blocks. This fusion simplifies the architecture while improving its ability to handle various sequence modeling tasks, including applications in language processing~\cite{mambamoe1,mambamoe2,wang2024mambabyte}, computer vision~\cite{mambaViT1,mambaViT2, mambaVideo}, medical imaging~\cite{mambaMedical7,mambaMedical5,mambaMedical8,mambaMedical4,mambaMedical3,mambaMedical2, mambaMedical1}, and graphs~\cite{mambaGraph1,mambaGraph2}. 
This fast adoption of SSMs and Mamba models underscores the need for reliable explanations of their predictions. 

\paragraph*{Explainable AI and SSMs.} In efforts to explain Mamba models, \cite{paulo2024doestransformerinterpretabilitytransfer} analyzed if the interpretability tools originally designed for Transformers can also be effectively applied to architectures such as Mamba. In this context, \citet{ali2024hidden} and \citet{zimerman2024unifiedimplicitattentionformulation} recently proposed viewing the internal computations of Mamba models as an attention mechanism. This approach builds upon previous works that use attention signal as explanation, including Attention Rollout \cite{abnar-zuidema-2020-quantifying} and variants thereof \cite{chefer2021transformer, Chefer_2021_ICCV}.
While these approaches can provide some insight, they inherit the limitations of using attention as an explanation \cite{wiegreffe-pinter-2019-attention, jain-wallace-2019-attention}, including their inability to capture potential misalignment between tokens and attention scores, and the limited performance in empirical faithfulness evaluations. Alternative Explainable AI methods, not yet applied to Mamba models but in principle applicable to any model, include techniques using input perturbations \cite{zeiler2014visualizing, zintgraf2017visualizing, fongperturbation2017} or leveraging gradient information \cite{baehrens10a, GI, intgrad, smoothgrad, gradcam}. Despite their wide applicability, these models have certain drawbacks, such as requiring multiple function evaluations for a single explanation or being susceptible to gradient noise, resulting in subpar performance, as our benchmark experiment will demonstrate. 
Alternatively, deriving tailored approaches that reflect the underlying model structure, has shown to be a promising direction in developing better attribution methods based on gradient analysis of the prediction function \cite{Arras2019, Eberle2022, schnake2022higher, DBLP:conf/icml/AliSEMMW22}. In the Layer-wise Relevance Propagation framework, this necessitates suitable propagation rules, which are currently lacking for the Mamba architecture. 
To tackle these challenges, we introduce MambaLRP as an efficient solution for the computation of reliable and faithful explanations that are theoretically grounded in the axiom of relevance conservation.

\section{Background}
\label{section:background}
Before delving into the details of our proposed method, we begin with a brief overview of the selective SSM architecture, followed by an introduction to the LRP framework. 
\paragraph{Selective SSMs (S6)}
An important component within the Mamba \cite{mamba} architecture is the selective SSM. It is characterized by parameters, $\bar{A}$, $\bar{B}$, and $C$, and transforms a given input sequence $(x_t)_{t=1}^T$
into an output sequence of the same size $(y_t)_{t=1}^T$ via the following equations:
\begin{align}
   h_t &= \bar{A_t}h_{t-1} + \bar{B_t}x_t
   \label{eq:hidden_state_update}\\
   y_t &= C_t h_t
   \label{eq:ssm_output}
\end{align}
where the initial state $h_0=0$. What distinguishes the selective SSM from the original SSM (S4) \cite{gu2022efficiently} is that the evolution parameter, $\bar{A_t}$, and projection parameters, $\bar{B_t}$ and $C_t$, are functions of the input $x_t$. This enables dynamic adaptation of the SSM's parameters based on input. This dynamicity facilitates focusing on relevant information while ignoring irrelevant details when processing a sequence. 

\paragraph{Layer-wise Relevance Propagation}
Layer-wise Relevance Propagation (LRP) \citep{bach-plos15} is an Explainable AI method that attributes the model's output to the input features through a single backward pass. This backward pass is specifically designed to identify neurons relevant to the prediction. LRP assigns relevance scores to neurons in a given layer and then propagates these scores to neurons in the preceding layer. The process continues layer by layer, starting from the network's output and terminating once the input features are reached. The LRP backward pass relies on an axiom called `conservation' requiring that relevance scores are preserved across layers, avoiding to artificially amplify or suppress contributions. For example, let $x$ and $y$ be the input and output of some layer, respectively, and let $\mathcal{R}(x)$ and $\mathcal{R}(y)$ represent the sum of relevance scores in the respective layers. The conservation axiom requires that $\mathcal{R}(x) = \mathcal{R}(y)$ holds true. 
\section{LRP for Mamba}
\label{section:mambalrp}
In this work, we bring explainability, particularly LRP, to Mamba models, following the conceptual design steps, shown in Fig.~\ref{fig:overview}. 
We start by applying a basic LRP procedure, specifically one corresponding to Gradient$\,\times\,$Input (GI), to the Mamba architecture. This serves as an effective initial step for identifying layers where certain desirable explanation properties, like relevance conservation, are violated. We analyze different layers of the Mamba architecture, derive relevance propagation equations and test the fulfillment of the conservation property. Our analysis reveals three components in the Mamba architecture where conservation breaks: the SiLU activation function, the selective SSM, and the multiplicative gating of the SSM's output. Leveraging the analysis above, we propose novel relevance propagation strategies for these three components, which lead to a robust, faithful and computationally efficient explanation approach, called MambaLRP.
\subsection{Relevance propagation in SiLU layers}
\label{section:silu} 
We start by examining the relevance propagation through Mamba's SiLU activation functions.  This function is represented by the equation $y = x \cdot \sigma(x)$, where $\sigma$ denotes the logistic sigmoid function. 
\begin{proposition}
Applying the standard gradient propagation equations yields the following result, which relates the relevance values before and after the activation layer:
\begin{align}
\label{eq:silu}
\xbrace{\dd{f}{x} x}{\R(x)}
= \ybrace{\dd{f}{y} y}{\R(y)}
+ \ebrace{\dd{f}{y} \cdot \sigma'(x) \cdot x^2}{\varepsilon}
\end{align}
\end{proposition}
The derivation for Eq.~\ref{eq:silu} can be found in Appendix \ref{section:silu-derivation}. We observe that the conservation property, \ie $\R(x) = \R(y)$, is violated whenever the residual term $\varepsilon$ is non-zero. We propose to restore the conservation property in the relevance propagation pass by locally expanding the SiLU activation function as:
\begin{align}
    y = x \cdot [\sigma(x)]_\text{cst.}
    \label{eq:siludetach}
\end{align}
where $[\cdot]_\text{cst.}$ treats the given quantity as constant. This can be implemented e.g.\ in PyTorch using the \texttt{.detach()} function. Repeating the derivation above with this modification yields the desired conservation property, $\R(x) = \R(y)$.  The explicit LRP rule associated to this LRP procedure is provided in Appendix \ref{sec:lrp_rules}.

\subsection{Relevance propagation in selective SSMs (S6)}
\label{section:ssm}
\begin{wrapfigure}{r}{0.5\linewidth}
\vspace{-3pt}
\centering
\includegraphics[width=0.35\textwidth]{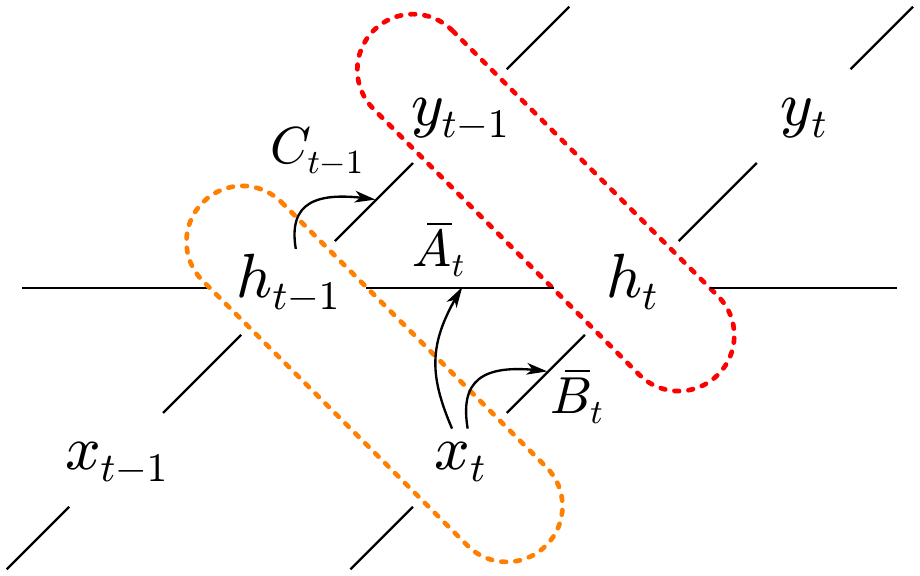}
\captionof{figure}{Unfolded view of SSM, highlighting two subsets of nodes, the relevance of which should be conserved throughout relevance propagation.}
\label{fig:ssm-unfolded}
\vspace{-8pt}
\end{wrapfigure}
The most crucial non-linear component of the Mamba architecture is its selective SSM component. It is designed to selectively retain or discard information throughout the sequence by adjusting its parameters based on the input, enabling dynamic adaptation to each token. To facilitate the analysis, we introduce an inconsequential modification to the original SSM by connecting $C_t$ to $h_t$ instead of $x_t$. To do so, we can redefine $\bar{A_t}$, $\bar{B_t}$, and $C_t$ matrices as $\mathrm{blockdiag}(\bar{A}_t \,,\,0)$, $(\bar{B}_t\,,\, I)$, and $(C_t \,|\, 0)$ respectively, such that $x_t$ becomes part of the state $h_t$ without altering the overall functionality of the SSM. 

The unfolded SSM, with the aforementioned modification, is illustrated in Fig.~\ref{fig:ssm-unfolded}. The complex relevance propagation procedure in the SSM component can be further simplified by considering two groups of units, illustrated in red and orange in Fig.~\ref{fig:ssm-unfolded}. In these two groups, there are no connections within units of the same group, all the relevance propagation signals from the first group are directed towards the second group, and the second group receives no further incoming relevance propagation signal. With these properties, these two groups should, according to the principle of conservation, receive the same relevance scores.
\begin{proposition}
    Defining $\theta_t = (\bar A_t,\bar B_t,C_{t-1})$, and working out the propagation equations between these two groups yields the following relation:
    \begin{align}
        \xbrace{\dd{f}{x_t} x_t + \dd{f}{h_{t-1}} h_{t-1}}{\mathcal{R}(x_t) + \mathcal{R}(h_{t-1})}
        = \ybrace{\dd{f}{h_t} h_t + \dd{f}{y_{t-1}} y_{t-1}}{\mathcal{R}(h_t) + \mathcal{R}(y_{t-1})}
        + \ebrace{\dd{f}{\theta_t} \dd{\theta_t}{x_{t}} x_{t} + \dd{f}{\theta_t} \dd{\theta_t}{h_{t-1}} h_{t-1}}{\varepsilon}
        \label{eq:dissipation-ssm}
    \end{align}
\end{proposition}
The derivation for Eq.~\ref{eq:dissipation-ssm} can be found in Appendix \ref{section:ssm-derivation}. We note that the residual term $\epsilon$, which is typically non-zero, violates conservation. Specifically, conservation fails due to the dependence of $\theta$ on the input. We propose to rewrite the state-space model at each step in a way that the parameters $\theta_t$ appear constant, \ieending:
\begin{align}
   h_t &= [\bar{A_t}]_\text{cst.} h_{t-1} + [\bar{B_t}]_\text{cst.}x_t   \label{eq:ssmdetach1}\\
   y_t &= [C_t]_\text{cst.} h_t
   \label{eq:ssmdetach2}
\end{align}
These equations can also be interpreted as viewing the selective SSM as a localized non-selective, \ie standard, SSM. With this modification, conservation holds between the two groups, \ie  $\mathcal{R}(x_t) + \mathcal{R}(h_{t-1}) = \mathcal{R}(h_t) + \mathcal{R}(y_{t-1})$. By repeating the argument for each time step, conservation is also maintained between the input and output of the whole SSM component. Explicit LRP rules are provided in Appendix \ref{sec:lrp_rules}.

\subsection{Relevance propagation in multiplicative gates}
\label{section:product}
In each block within the Mamba architecture, the SSM's output is multiplied by an input-dependent gate. In other words, $y = z_A \cdot z_B$, where $z_A = \text{SSM}(x)$ and $z_B = \text{SiLU}(\text{Linear}(x))$. Assume that the locally linear expansions introduced in Sections \ref{section:silu} and \ref{section:ssm} are applied to the SSM components and SiLU activation functions, the mapping from $x$ to $y$ becomes quadratic. 
\begin{proposition}
Applying the standard gradient propagation equations establishes the following relation between the relevance values before and after the gating operation:
\begin{align}
\xbrace{\dd{f}{x} x}{\R(x)}
= \ybrace{\dd{f}{y} y}{\R(y)}
+ \ebrace{\dd{f}{y} y}{\varepsilon}
\label{eq:gate}
\end{align}
\end{proposition}
The derivation for Eq.~\ref{eq:gate} and explicit LRP rules can be found in Appendix \ref{section:product-derivation} and Appendix~\ref{sec:lrp_rules}, respectively. In this equation, we observe a spurious doubling of relevance in the backward pass. This can be addressed by treating half of the output as constant:
\begin{align}
y = 0.5 \cdot (z_A \cdot z_B) + 0.5 \cdot [z_A \cdot z_B]_{\text{cst.}}
\label{eq:half}
\end{align}
As for the previous examples, this ensures the conservation property $\R(x) = \R(y)$. An alternative would have been to make $y$ linear by detaching only one of the terms in the product, as done for the SiLU activation or the SSM component. However, the strategy of Eq.~\ref{eq:half} better maintains the directionality given by the gradient. We further compare these alternatives in an ablation study presented in Appendix \ref{sec:ablation_multiplicative_gate}, demonstrating empirically that our proposed approach performs better.

\subsection{Additional modifications and summary}
\label{section: additional_modifications}
The propagation strategies developed for the Mamba-specific components complement previously proposed approaches for other layers, including propagation through RMSNorm layers \cite{DBLP:conf/icml/AliSEMMW22} and convolution layers  via robust LRP-$\gamma$ rules \citep{montavon2019layer, DOMBROWSKI2022108194} and their generalized variants. A summary of these additional enhancements is provided in Appendix~\ref{sec:mambalrp_details}. Furthermore, our proposed propagation rules are generally applicable to other models that utilize similar components, such as multiplicative gates in recent architectures \cite{qin2023hierarchically, peng-etal-2023-rwkv, ma2023mega, mamba2_dao24a}.

A straightforward implementation of the propagation rules can be achieved by computing MambaLRP via Gradient $\times$ Input, where the gradient computations are modified to align with the proposed rules. The procedure consists of two main steps:
\begin{enumerate}
\item Perform the detach operations of Eqs.\ \eqref{eq:siludetach}, \eqref{eq:ssmdetach1}, \eqref{eq:ssmdetach2}, and \eqref{eq:half} (as well as similar operations for RMSNorm and convolutions).
\item Retrieve MambaLRP explanations by computing Gradient$\,\times\,$Input on the detached model.
\end{enumerate}

\section{Experiments}
\label{sec:experiments}
To evaluate our proposed approach, we benchmark its effectiveness against various methods previously proposed in the literature for interpreting neural networks. We empirically evaluate our proposed methodology using Mamba-130M, Mamba-1.4B, and Mamba-2.8B language models \cite{mamba}, which are trained on diverse text datasets. The training details can be found in Appendix~\ref{sec:models_and_datasets}. For the vision experiments, we use the Vim-S model \cite{mambaViT2}. Moreover, we perform several ablation studies to further investigate our proposed method.
\paragraph{Datasets}
In this study, we perform experiments on four text classification datasets, namely SST-2 \citep{sst}, Medical BIOS \citep{medical_bios}, Emotion \citep{saravia-etal-2018-carer}, and SNLI \citep{snli}. The SST-2 dataset encompasses around 70K English movie reviews, categorized into binary classes, representing positive and negative sentiments.  The Medical BIOS dataset consists of short biographies (10K) with five specific medical occupations as targets. The SNLI corpus (version 1.0) comprises 570k English sentence pairs, with the labels entailment, contradiction, and neutral, used for the  natural language inference (NLI) task. The Emotion dataset (20K) is a collection of English tweets, each labeled with one of six basic emotions. For the vision experiments, we use ImageNet dataset \citep{imagenet} with 1.3M images and 1K classes.
\paragraph{Baseline methods}
We compare our proposed method with several gradient-based, model-agnostic explanation techniques: Gradient$\,\times\,$Input (GI) \citep{baehrens10a, GI}, SmoothGrad \cite{smoothgrad}, and Integrated Gradients \cite{intgrad}. Furthermore, we evaluate the performance of our proposed method against a naive implementation of LRP, \ie LRP (LN-rule), where the LRP-0 rule is used in all linear and convolution layers, along with the LN-rule \cite{DBLP:conf/icml/AliSEMMW22} in normalization layers. 

We further compare the performance of our proposed method with two attention-based approaches, Attention Rollout (AttnRoll) and MambaAttr \citep{ali2024hidden}, which are recently proposed for Mamba models. Both methods are extensions of techniques originally developed for Transformer models: Attention Rollout \cite{abnar-zuidema-2020-quantifying} and Gradient$\times$Attention Rollout \cite{chefer2021transformer}.

\subsection{Conservation property}
\begin{figure}[!htb]
    \centering
    \includegraphics[width=0.8\textwidth]{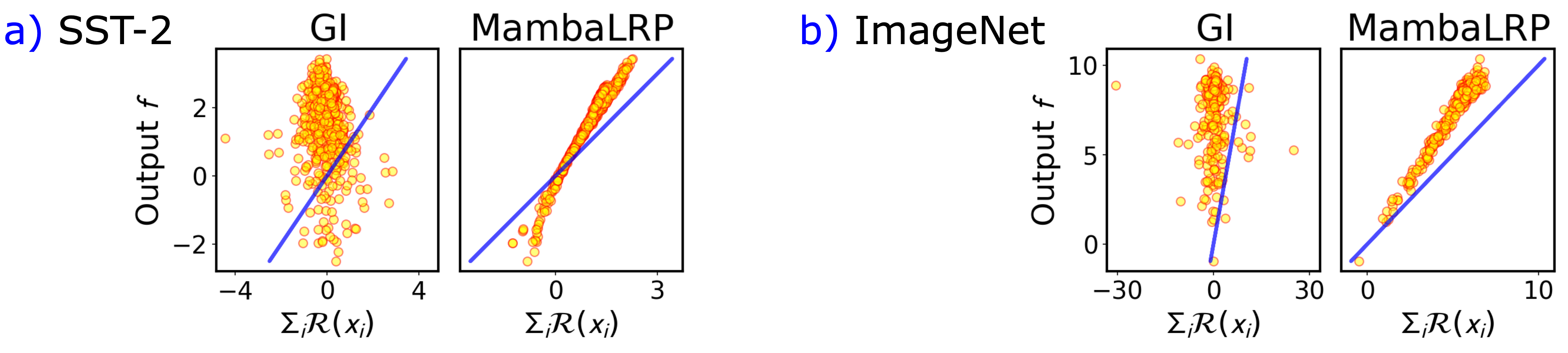}
    \caption{Conservation property. The x-axis represents the sum of relevance scores across the input features and the y-axis shows the network's output score. Each point corresponds to one example and its proximity to the blue identity line indicates the extent to which conservation is preserved, with closer alignment suggesting improved conservation.}
    \label{fig:conservation}
    \vspace{-8pt}
\end{figure}

To verify the fulfillment of the conservation property, on which our method is based, we compare the network's output score with the sum of relevance scores attributed to the input features, for both the GI baseline and the proposed MambaLRP. The analysis is performed for Mamba-130M and Vim-S models trained on the SST-2 and ImageNet datasets, respectively. Full conservation is achieved if the output score equals the sum of relevance, as indicated by the blue line in Fig.~\ref{fig:conservation}. Our results show that conservation is severely violated by the GI baseline, and is addressed to a large extent by MambaLRP. Residual lack of conservation is due to the presence of biases in linear and convolution layers, which are typically non-attributable.
\subsection{Qualitative evaluation}
\label{sec:qualitative_results}
\begin{figure}[!htb]
    \centering
    \includegraphics[width=0.9\textwidth]{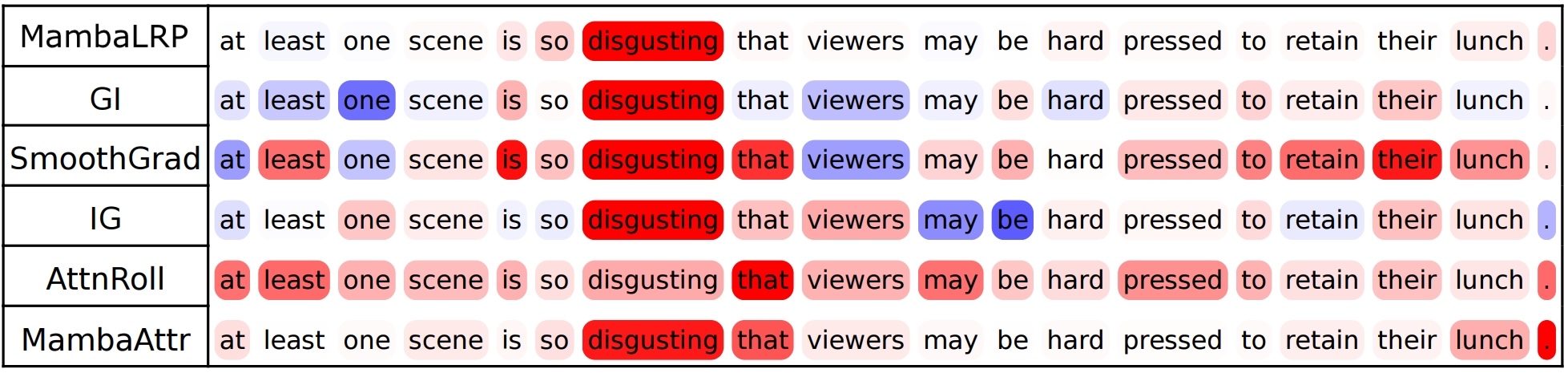}
    \caption{Explanations generated for a sentence of the SST-2 dataset. Shades of red represent words that positively influence the model's prediction. Conversely,  shades of blue reflect negative contributions. The heatmaps of attention-based methods are constrained to non-negative values.}
    \label{fig:sst_qualitative}
    \vspace{-2pt}
\end{figure}

\begin{figure}[!htb]
    \centering
    \includegraphics[width=0.8\textwidth]{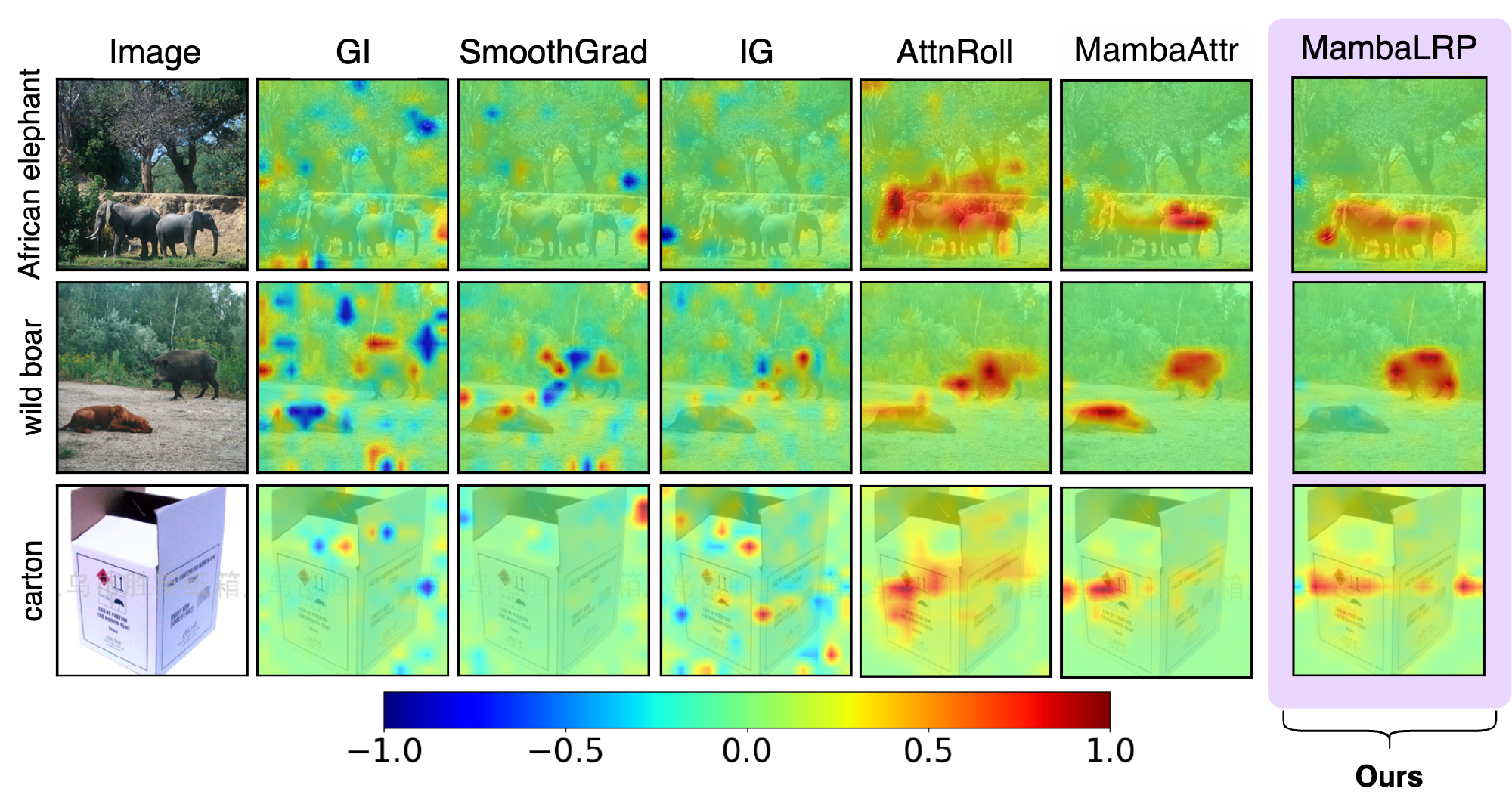}
    \caption{Explanations produced by different explanation methods for images of the ImageNet dataset. AttnRoll and MambaAttr are limited to non-negative heatmap values.}
    \label{fig:imagenet_qualitative}
    \vspace{-12pt}
\end{figure}

In this section, we qualitatively examine the explanations produced by various explanation methods for Mamba-130M and Vim-S models. Fig.~\ref{fig:sst_qualitative} illustrates the explanations generated to interpret the Mamba-130M model's prediction on a sentence from the SST-2 dataset with negative sentiment. We note that all of the explanation methods attribute positive scores to the word  `disgusting', which appears reasonable given the negative sentiment label. However, it is notable that the explanation generated by MambaLRP is more sparse and focuses particularly on the terms `so' and `disgusting'. In contrast, the explanations produced by the gradient-based methods and AttnRoll appear to be quite noisy. 
Furthermore, we show the explanations produced to interpret the Vim-S model's predictions on images of the ImageNet dataset in Fig.~\ref{fig:imagenet_qualitative}. Purely gradient-based explanations tend to identify unspecific noisy features, while both attention-based approaches, AttnRoll and MambaAttr, are more effective at highlighting significant features. Among these methods, MambaLRP stands out for its ability to generate explanations that are particularly focused on key features used by the model to make a prediction. Take, for instance, the first image classified under the `African elephant' category. We can see that the explanation generated by MambaLRP not only includes all occurrences of the `African elephant' object but also highlights its distinctive features, such as the tusks. In the second image labeled `wild boar', despite the presence of multiple objects in the image, MambaLRP's explanation remains focused on the `wild boar' object, disregarding other objects. Moreover, in the third instance, MambaLRP uncovers a spurious correlation, the presence of a watermark in Chinese, influencing the model's prediction, a subtlety overlooked or not fully represented by other methods. Further qualitative results can be found in Appendix~\ref{sec:additional_qualitative}.
\subsection{Quantitative evaluation}
\label{sec:quantitative_eval}
To quantitatively evaluate the faithfulness of explanation methods, we employ an input perturbation approach based on ranking input features by their importance \cite{Samek2015EvaluatingTV}, which can be done using either a Most Relevant First (MoRF) or Least Relevant First (LeRF) strategy. Ranked features are iteratively perturbed through a process known as \textit{flipping}. We monitor the resulting changes in the output logit, $f_c$, for the predicted class $c$, and compute the area under the perturbation curve. The areas under the curves for LeRF and MoRF strategies are denoted by $A^F_\mathrm{LeRF}$ and $A^F_\mathrm{MoRF}$, respectively. 
In contrast, the \textit{insertion} method starts with a fully perturbed input and progressively restores important features. The areas under the curves for this method are indicated by $A^I_\mathrm{MoRF}$ and $A^I_\mathrm{LeRF}$, for the MoRF and LeRF strategies, respectively. 
A reliable explanation method is characterized by low values of $A^F_\mathrm{MoRF}$ or $A^I_\mathrm{LeRF}$, and large values of $A^F_\mathrm{LeRF}$ or $A^I_\mathrm{MoRF}$. 
In an effort to minimize the introduction of out-of-distribution manipulations, the recent study by \citet{blücher2024decoupling} advocates for harnessing both insights to derive a more resilient metric. Therefore, we follow the same strategy as \cite{blücher2024decoupling, achtibat2024attnlrp} to evaluate explanation methods. The evaluation metrics are defined as $\Delta A^F = A^F_\mathrm{LeRF} - A^F_\mathrm{MoRF}$ and $\Delta A^I = A^F_\mathrm{MoRF} - A^F_\mathrm{LeRF}$. For both metrics, a higher score is preferable, as it signifies a more accurate and reliable explanation method.

The outcomes of this analysis are represented in Table~\ref{tab:flipping}. MambaLRP consistently achieves highest faithfulness scores in comparison to other baseline methods. We observe that GI struggles with noisy attributions, leading to low faithfulness scores. However, methods like Integrated Gradients and MambaAttr have shown improvements in this regard. We note that LRP (LN-rule) outperforms most methods across the majority of the text classification tasks. Nevertheless, its performance is notably inferior compared to MambaLRP. Overall, we observe that MambaLRP significantly outperforms all other methods by a substantial margin. In both vision and NLP experiments, attention-based methods have shown superior performance compared to the purely gradient-based approaches. 
\begin{table}[!htb]
\centering
\captionsetup{skip=0.5\baselineskip}
\caption{Evaluating explanation methods. Higher scores $\Delta{A}^F$ indicate more faithful explanations.}
\label{tab:flipping}
\resizebox{\textwidth}{!}{
\small
\addtolength{\tabcolsep}{-0.75em}
\begin{tabular}{lccccccccccccc}
\toprule
\multirow{2}{*}{\textbf{Methods}} &
\multicolumn{3}{c}{\textbf{SST-2}} &
\multicolumn{3}{c}{\textbf{{Med-BIOS}}}&
\multicolumn{3}{c}{\textbf{{SNLI}}} &
\multicolumn{3}{c}{\textbf{{Emotion}}} &
\multicolumn{1}{c}{\textbf{{ImageNet}}} \\
\cmidrule(rl){2-4} 
\cmidrule(rl){5-7} 
\cmidrule(rl){8-10}
\cmidrule(rl){11-13}
\cmidrule(rl){14-14}
&
{{\parbox{1.2cm}{\centering Mamba \\ 130M}}} & {{\parbox{1.2cm}{\centering Mamba \\ 1.4B}}} & {{\parbox{1.2cm}{\centering Mamba \\ 2.8B}}} & {{\parbox{1.2cm}{\centering Mamba \\ 130M}}} & {{\parbox{1.2cm}{\centering Mamba \\ 1.4B}}} & {{\parbox{1.2cm}{\centering Mamba \\ 2.8B}}} & {{\parbox{1.2cm}{\centering Mamba \\ 130M}}} & {{\parbox{1.2cm}{\centering Mamba \\ 1.4B}}} & {{\parbox{1.2cm}{\centering Mamba \\ 2.8B}}} & {{\parbox{1.2cm}{\centering Mamba \\ 130M}}} & {{\parbox{1.2cm}{\centering Mamba \\ 1.4B}}} & {{\parbox{1.2cm}{\centering Mamba \\ 2.8B}}} & {Vim-S} \\
\midrule
\centering
Random & -0.012 & -0.106 & 0.007 & 0.044 & -0.014 & -0.037 & 0.010 & 0.002 & 0.000 & -0.001 & 0.000 & 0.000 & -0.001\\
\hline & \\[-1.5ex]
GI \citep{GI} & 0.078 & -0.106 & -0.043 & 0.200 & -0.634 & -1.434 & -0.039 & -0.039 & 0.083 & -0.787 & -0.409 & -1.533 & -0.018\\
SmoothGrad \citep{smoothgrad} & 1.377 & -0.383 & -0.675 & 1.661 & -2.300 & -1.908 & 0.486 & -0.687 & -0.747 & 1.808 & -1.852 & -4.228 & 0.209\\ 
IG \citep{intgrad} & 0.857 & 0.216 & 0.322 & 1.296 & 1.065 & 1.937 & 0.453 & 0.218 & 0.331 & 1.808 & 2.010 & 4.314 & 1.217\\
\hline & \\[-1.5ex]
AttnRoll \citep{ali2024hidden} & 0.657 & 0.431 & 0.452 & 2.228 & 1.076 & 2.241 & 0.242 & 0.371 & 0.292 & 0.389 & 1.483 & 0.530 & 2.427\\
MambaAttr \citep{ali2024hidden} & 1.190 & 0.626 & 0.341 & 3.126 & 3.006 & 5.326 & 0.513 & 0.554 & 0.343 & 2.003 & 4.706 & 3.849 & 2.676\\
\hline & \\[-1.5ex]
LRP (LN-rule, \citep{DBLP:conf/icml/AliSEMMW22}) & 0.877 & 0.961 & 0.820 & 2.217 & 3.456 & 5.305 & 0.673 & 0.656 & 0.731 & 3.079 & 5.199 & 5.094 & 2.548 \\
\rowcolor{cyan!10} MambaLRP (ours) & \textbf{1.978} & \textbf{1.248} & \textbf{1.157} & \textbf{3.906} & \textbf{4.234} & \textbf{7.083} & \textbf{0.989} & \textbf{0.897} & \textbf{0.899} & \textbf{3.523} & \textbf{5.397} & \textbf{5.637} & \textbf{4.715} \\
\bottomrule
\end{tabular}
}
\vspace{-8pt}
\end{table}
\paragraph{Runtime comparison} We report the runtimes of MambaLRP along with other methods used in this study in Appendix~\ref{sec:runtime}. As shown in Table~\ref{tab:app_runtime}, our method's runtime is comparable to GI and can be implemented via a single forward and backward pass. Since approaches like Integrated Gradients require multiple function evaluations, their runtimes are considerably higher than MambaLRP.
\paragraph*{Ablation study} In Section \ref{section:mambalrp}, we proposed techniques for handling different non-linear components within the Mamba architecture. This ablation study aims to assess the significance of each technique by testing the effect of their exclusion on faithfulness. Table~\ref{tab:mambalrp_components} shows that all three modifications are essential for achieving competitive explanation performance, with our proposed method for handling the SSM component being the most critical. Further experiments, comparing different strategies for handling the Mamba block's multiplicative gate, are detailed in Appendix~\ref{sec:ablation_multiplicative_gate}.

\begin{minipage}[t]{0.45\textwidth}
    \centering
    \captionof{table}{Analyzing the impact of ablating the three proposed propagation rules on $\Delta{A}^F$ for the components in MambaLRP.}
    \resizebox{0.9\textwidth}{!}{
    \begin{tabular}{ccc|cc}
         \toprule
         \textbf{SiLU} & \textbf{SSM} & \textbf{Gate} & \textbf{SST-2} & \textbf{ImageNet} \\
         \toprule
         \cmark & \xmark & \cmark & 0.577 & 0.144 \\
         \cmark & \cmark & \xmark & 1.721 & 4.022 \\
         \xmark & \cmark & \cmark & 1.943 & 4.618 \\
         \rowcolor{cyan!10} \cmark & \cmark & \cmark & \textbf{1.978} & \textbf{4.715} \\
         \bottomrule  
    \end{tabular}
    }
    \label{tab:mambalrp_components}
\end{minipage}
\hfill
\begin{minipage}[t]{0.5\textwidth}
\centering
\captionof{table}{Frequency of gendered words in explanations for `Nurse' and `Surgeon' classes of the Medical BIOS dataset across language models.}
\label{tab:gender_bias_bios}
\resizebox{0.65\textwidth}{!}{
\begin{tabular}{l|cc}
\toprule
{\rotatebox{0}{\textbf{Models}}} &
{\rotatebox{0}{\textbf{Surgeon}}} &
{\rotatebox{0}{\textbf{Nurse}}} \\
\midrule
GPT2-base & 0.14 & 0.24\\
T5-base & 0.10 & 0.11\\
RoBERTa-base & 0.01 & 0.06\\
Mamba-130M & 0.009 & 0.058\\
Mamba-1.4B & 0.001 & 0.042 \\
\bottomrule
\end{tabular}
}
\vspace{-8pt}
\end{minipage}
\section{Use cases}
\label{sec:use_cases}
\paragraph{Uncovering gender bias in Mamba.}
Explanation methods serve as tools to uncover biases in pretrained vision and language models. Using our proposed method, we examine Mamba-130M and Mamba-1.4B models, trained on the Medical BIOS dataset, to investigate the potential presence of gender biases. Following the methodology in \citep{medical_bios}, we use MambaLRP to identify the top-5 tokens of highest importance and to quantify the prevalence of gendered words within these tokens. We find that the model exhibits a pronounced preference for female-gendered words in the `Nurse' class (e.g. the proportion of gender-specific words is 0.058 for females, compared to 0.0 for males in Mamba-130M.). We also compare the results of our analysis with those achieved for the GPT2-base, T5-base, and RoBERTa-base models as mentioned in \citep{medical_bios}. As shown in Table \ref{tab:gender_bias_bios}, both Mamba models are less dependent on gendered tokens compared to GPT2-base, T5-base, and RoBERTa-base models, with the Mamba-1.4B model showing a further decrease in bias compared to the Mamba-130M, suggesting improvements in reducing gender bias with increased model size.

\begin{figure}[!t]
    \centering
    \includegraphics[width=1.\textwidth]{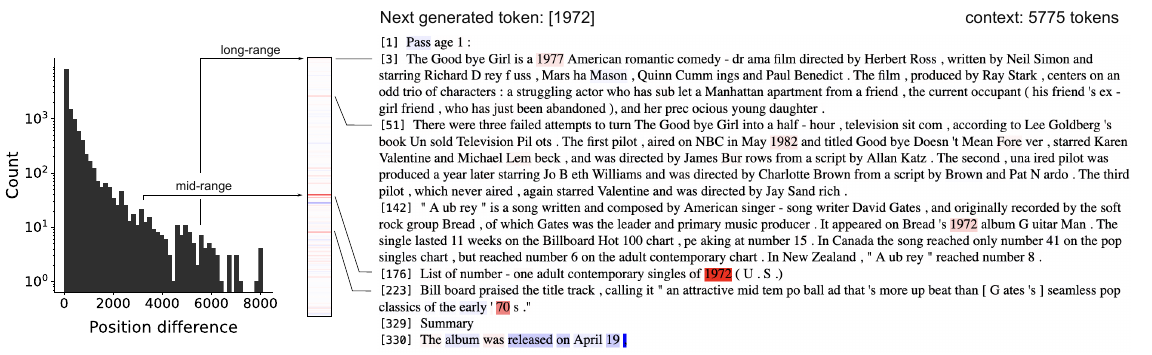}
    \caption{Analysis of the position of tokens relevant for next token generation. Left: Distribution of absolute position of the ten most relevant tokens for the prediction of the next word. Right: Long-range dependency between tokens of the input and the predicted next token (here: \textit{1972}). }
    \label{fig:lrd}
\end{figure}

\paragraph{Investigating long-range capabilities of Mamba.}
The ability of SSMs to model long-range dependencies is considered an important improvement over previous sequence models. In this use case, we analyze the extent to which the pretrained Mamba-130M model can use information from the entire context window.  We use the HotpotQA \cite{yang2018hotpotqa} subset from the LongBench dataset \cite{bai2023longbench}, designed to test long context understanding. After selecting all 127 instances, containing sequences up to 8192 tokens, we prompt the model to summarize the full paragraph by generating ten additional tokens. Fig.~\ref{fig:lrd} shows the distribution of the positional difference between a relevant token and the currently generated token. While we observe a pronounced pattern of attributing to the last few tokens, as seen in prior language generation studies \cite{yin-neubig-2022-interpreting,sarti-etal-2023-inseq}, the extracted explanations also identified relevant tokens across the entire context window, as presented for one example in Fig.~\ref{fig:lrd} (right). This suggests that the model is indeed capable of retrieving long-range dependencies. We clearly see that in order to complete the sentence and assign a year to the album release date, the model analyzes previous occurrences of chronological information and MambaLRP identifies evidence supporting the decision for the date being `1972' as relevant. Our analysis demonstrates the previously speculated long-range abilities of the Mamba architecture \cite{mamba}, which we further explore in a comparison to Transformers in Appendix~\ref{sec:lrd_transformers}.

\begin{figure}[!t]
    \centering\includegraphics[width=\textwidth]{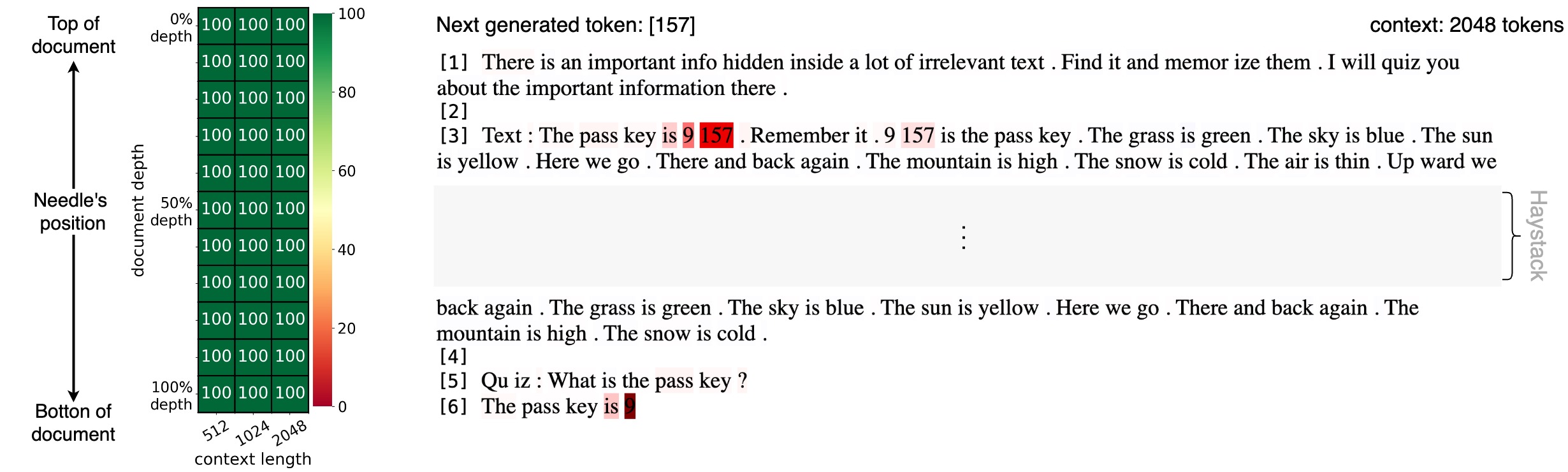}
    \caption{Explanation-based retrieval accuracy in the needle-in-a-haystack test verifying model reliance on relevant features for different context lengths.}
    \label{fig:neddle}
    \vspace{-8pt}
\end{figure}
\paragraph{Needle-in-a-haystack test.} 
\label{sec:needle}
To assess the model's ability in retrieving relevant pieces of information from a broader context, we perform the needle-in-a-haystack test \cite{needle_1}. Our test involves extracting a single passkey (the `needle') from a collection of repeated noise sentences (the `haystack'), as described in \cite{hsieh2024ruler}. We run this test at eleven different document depths with three different context lengths. We use an instruction-finetuned Mamba-2.8B model in this experiment.
To analyze the performance of the model, we introduce the explanation-based retrieval accuracy (XRA) metric. In this approach, we first identify the positions of the top-K relevant tokens by MambaLRP, and then, calculate the accuracy by comparing those positions to the needle's position. As shown in Fig.~\ref{fig:neddle}, MambaLRP accurately captures the information used by the model to retrieve the needle. In this case, the model could accurately retrieve the needle based on relevant information within the text. However, in more realistic and complex scenarios, the model may depend on irrelevant data yet still generate the correct token. This issue can be analyzed using XRA but cannot be evaluated by conventional retrieval accuracy metrics. Such cases and also further details about this experiment are shown in Appendix~\ref{sec:additional_use_case}.
\section{Discussion and conclusion}
\label{sec:conclusion}
Mamba models have emerged as an efficient alternative to Transformers. However, there are limited works addressing their interpretability \cite{ali2024hidden, zimerman2024unifiedimplicitattentionformulation}.
To address this issue, we proposed MambaLRP within the LRP framework, specifically tailored to the Mamba architecture and built upon the relevance conservation principle. Our evaluations across various models and datasets confirmed that MambaLRP  adheres to the conservation property and provides faithful explanations that outperform other methods while being more computationally efficient.  
Moreover, we demonstrated how MambaLRP can help users debug state-of-the-art vision and language models while building trust in their predictions through various use cases.
Future research can explore its potential across a broader range of applications and Mamba architectures, providing reliable insights into sequence models.  

\paragraph{Limitations} As a propagation-based explanation method, MambaLRP requires storing activations and gradients, leading to memory usage that depends on the model architecture and input sequence length. To reduce memory consumption, techniques such as gradient checkpointing can be utilized, which are applicable to other gradient-based methods as well. However, a limitation of these methods, including MambaLRP, is the potential inaccessibility of gradient information due to proprietary constraints. In such cases, approximating gradient information may offer a viable solution.

\section*{Acknowledgments}
This work was funded by the German Ministry for
Education and Research (refs. 01IS14013A-E, 01GQ1115, 01GQ0850, 01IS18025A, 031L0207D, 01IS18037A). K.R.M.\ was partly supported by the Institute of Information \& Communications Technology Planning \& Evaluation (IITP) grants funded by the Korea government (MSIT) (No. 2019-0-00079, Artificial Intelligence Graduate School Program, Korea University and No. 2022-0-00984, Development of Artificial Intelligence Technology for Personalized Plug-and-Play Explanation and Verification of Explanation).

\medskip
{
\small
\bibliographystyle{abbrvnat}
\bibliography{references}
}

\newpage
\appendix

\section{Proofs}
In the following, we provide derivations for the conservation analysis performed in Section~\ref{section:mambalrp}.
\subsection{Derivations for SiLU}
\label{section:silu-derivation}
We first consider the SiLU activation function. As mentioned in Section~\ref{section:silu}, this function is represented by the equation $y = x \cdot \sigma(x)$, with $\sigma$ being the logistic sigmoid function. By applying the standard gradient propagation equations, we get the conservation equation:

\begin{align}
\begin{split}
\Xbrace{\dd{f}{x} x}{\R(x)}
&= \dd{f}{y} \dd{y}{x} x\\
&= \dd{f}{y} \cdot (\sigma(x) + x \sigma'(x)) \cdot x\\
&= \dd{f}{y} \cdot \sigma(x) \cdot x + \dd{f}{y} \cdot x \sigma'(x) \cdot x\\
&= \ybrace{\dd{f}{y} y}{\R(y)} + 
\ebrace{\dd{f}{y} \cdot \sigma'(x) \cdot x^2}{\varepsilon}
\end{split}
\end{align}

\subsection{Derivations for selective SSM}
\label{section:ssm-derivation}
In Section \ref{section:ssm}, we introduced an inconsequential modification to the original selective SSM architecture by connecting the matrix $C_t$ to the state $h_t$ instead of the input $x_t$. The unfolded view of the SSM component with this modification is represented in Fig. \ref{fig:ssm-unfolded}. We can observe two subsets of nodes in this figure. The relevance scores of these two subsets should be equal if the conservation property holds. Computing the relevance propagation equation between these two groups, we obtain:

\begin{equation}
\begin{split}
\Xbrace{\dd{f}{x_t} x_t + \dd{f}{h_{t-1}} h_{t-1}}{\R(x_t) + \R(h_{t-1})}
&=
\Big(
  \dd{f}{h_t} \dd{^+ h_t}{x_t} x_t
+ \dd{f}{\theta_t} \dd{\theta_t}{x_t} x_t
\Big) \\
&\qquad
+ 
\Big(
  \dd{f}{y_{t-1}} \dd{^+ y_{t-1}}{h_{t-1}}h_{t-1}
+ \dd{f}{h_t} \dd{h_t}{h_{t-1}} h_{t-1}
+ \dd{f}{\theta_t} \dd{\theta_t}{h_{t-1}} h_{t-1}
\Big)
\\
&=
\Big(
  \dd{f}{h_t} \dd{^+ h_t}{x_t} x_t
+ \dd{f}{h_t} \dd{h_t}{h_{t-1}} h_{t-1}
\Big)
+
\Big(\dd{f}{y_{t-1}} \dd{^+ y_{t-1}}{h_{t-1}}h_{t-1}\Big) \\
&\qquad
+
\Big(
\dd{f}{\theta_t} \dd{\theta_t}{x_t} x_t
+ \dd{f}{\theta_t} \dd{\theta_t}{h_{t-1}} h_{t-1}
\Big)
\\
&= \ybrace{\dd{f}{h_t} h_t + \dd{f}{y_{t-1}} y_{t-1}}{\mathcal{R}(h_t) + \mathcal{R}(y_{t-1})} +  \ebrace{\dd{f}{\theta_t} \dd{\theta_t}{x_t} x_t + \dd{f}{\theta_t} \dd{\theta_t}{h_{t-1}} h_{t-1}}{\varepsilon}
\end{split}
\end{equation}

\subsection{Derivations for multiplicative gate}
\label{section:product-derivation}
Mamba is composed of several blocks. In each block, the selective SSM's output is multiplied by an input-dependent gate. In other words, $y=z_Az_B$ with $z_A=\text{SSM}(x)$ and $z_B=\text{SiLU}(\text{Linear}(x))$. By applying the standard gradient propagation equations, we get the conservation equation:

\begin{align}
\begin{split}
\Xbrace{\dd{f}{x} x}{\R(x)}
&= \dd{f}{y} \dd{y}{x} x\\
&= \dd{f}{y} \cdot \Big(\dd{z_A}{x} z_B + z_A \dd{z_B}{x} \Big) x\\
&= \dd{f}{y} \cdot \big(z_A z_B + z_A z_B \big)\\
&= \ybrace{\dd{f}{y} y}{\R(y)} + \ebrace{\dd{f}{y} y}{\varepsilon}
\end{split}
\end{align}
\section{Explicit propagation rules for MambaLRP}
\label{sec:lrp_rules}

Whereas MambaLRP is more easily implemented via the modified gradient-based approach described in the main paper, we provide below explicit relevance propagation equations for better comparability with other works. We refer to Sections \ref{section:background} and \ref{section:mambalrp} of the main paper for the definition of the notation.

\subsection{SiLU}
Explicit LRP rule for SiLU layers:
\begin{align}    
\mathcal{R}(x_i) 
    &= 
    \mathcal{R}(y_i)
\end{align}

\subsection{SSM}
Using the shortcut notations
$a_{ij} = [A_t(x_t)]_{ji}$,
$b_{ij} = [B_t(x_t)]_{ji}$ and
$c_{ij} = [C_{t-1}(h_{t-1})]_{ji}$,
we can write the propagation of relevance to the previous state space activations explicitly as:
\begin{align}
    \mathcal{R}(h_i^{(t-1)})
    &= \sum_j \frac{ h_i^{(t-1)} c_{ij}}{\sum_i  h_i^{(t-1)} c_{ij}} \mathcal{R}(y_j^{(t-1)})
    +
    \sum_j \frac{h_i^{(t-1)} a_{ij}}{\sum_i h_i^{(t-1)} a_{ij} + \sum_{i'} x_{i'}^{(t)} b_{i'j}} \mathcal{R}(h_j^{(t)})
\end{align}
and the propagation of relevance to the SSM input as:
\begin{align}
    \mathcal{R}(x_i^{(t)})
    &= \sum_j \frac{ x_i^{(t)} b_{ij}  } {\sum_{i} x_i^{(t)} b_{ij} + \sum_{i'} h_{i'}^{(t-1)} a_{i'j}} \mathcal{R}(h_j^{(t)})
\end{align}

\subsection{Multiplicative Gate}

Explicit LRP rule for the multiplicative gate:
\begin{align}
    \mathcal{R}([z_A]_i) = 0.5 \cdot \mathcal{R}(y_i)\\
    \mathcal{R}([z_B]_i) = 0.5 \cdot \mathcal{R}(y_i)
\end{align}
\section{Experimental details}
\label{sec:experimental_details}
In this section, we provide experimental details on our experiments that allow  reproducibility of our results.
\subsection{Models and datasets}
\label{sec:models_and_datasets}
For the NLP experiments, we fine-tuned all parameters of the pretrained Mamba-130M, Mamba-1.4B, and Mamba-2.8B models\footnote{\url{https://github.com/state-spaces/mamba}} on four text classification datasets: SST-2, SNLI, Medical BIOS, and Emotion. The data statistics can be seen in Table~\ref{tab:dataset_statistics}. For the vision experiments, we used the pretrained Vim-S model\footnote{\url{https://github.com/hustvl/Vim}}, trained on the ImageNet dataset.
\paragraph{Training details}
During training, we used a batch size of 32. To train the Mamba-1.4B and Mamba-2.8B models on the SNLI dataset, a batch size of 64 is used. We employed the \{EleutherAI/gpt-neox-20b\}\footnote{\url{https://github.com/EleutherAI/gpt-neox}} tokenizer. The models' parameters were optimized using AdamW optimizer with a learning rate set at $7e-5$. Additionally, we used a linear learning rate scheduler with an initial factor of 0.5. All models were trained for a maximum of 10 epochs. We employed early stopping and ended training as soon as the validation loss ceased to improve. The top-1 accuracies of the models on each dataset are detailed in Table~\ref{tab:model_accuracy}. 

\begin{minipage}[!ht]{0.5\textwidth}
    \captionsetup{skip=0.5\baselineskip}
    \centering
    \captionof{table}{The accuracies of Mamba-130M, Mamba-1.4B, and Mamba-2.8B models on the validation sets of four text classification datasets.}
    \small
    \begin{tabular}{l|ccc}
         \toprule
         \textbf{Dataset} & \textbf{130M} & \textbf{1.4B} & \textbf{2.8B} \\
         \midrule
          SST-2 &  91.97 & 94.15 & 94.26\\
          Med-BIOS & 89.10 & 90.30 & 90.50\\
          Emotion & 93.45 & 93.65 & 94.10\\
          SNLI & 89.57 & 91.05 & 91.14\\
         \bottomrule
    \end{tabular}
    \label{tab:model_accuracy}
\end{minipage}
\hfill
\begin{minipage}[!ht]{0.5\textwidth}
    \captionsetup{skip=0.5\baselineskip}
    \centering
    \captionof{table}{Data statistics.}
    \small
    \begin{tabular}{l|ccc}
         \toprule
         \textbf{Dataset} & \textbf{Train} & \textbf{Test} & \textbf{Validation}\\
         \midrule
         SST-2 & ~68K & ~2K & ~1K\\
         Med-BIOS & 8K & 1K & 1K\\
         Emotion & ~16K & 2K & 2K\\
         SNLI & ~550K & 10K & 10K\\
         ImageNet & ~1.3M & 50K & 100K\\
         \bottomrule
    \end{tabular}
    \label{tab:dataset_statistics}
    \vspace{-12pt}
\end{minipage}
\subsection{MambaLRP details}
\label{sec:mambalrp_details}
In this section, we begin by showing how MambaLRP can be implemented through the following algorithms. Then, we explain the generalized LRP-$\gamma$ rule,  provide details regarding hyperparameters used in our implementation, and outline the hyperparameter selection procedure.
\begin{tcolorbox} 
[colback=white!100, colframe=SkyBlue!60, width=\textwidth, 
righttitle=0.5cm, subtitle style={boxrule=0.5pt, 
colback=yellow!50!red!25!white}, title=\textcolor{black}{\textbf{\normalsize{Algorithm 1: }}{MambaLRP in SiLU activation layer}}]
    {
    \begin{algorithm}[H]
    \KwData{\textbf{Input:} $x$ \textcolor{ForestGreen}{(B, L, D)}}
        $z \leftarrow \text{Identity}(x)$ \\
        \textbf{return} $z \odot [\text{SiLU}(x) \oslash z]\textcolor{Plum}{\texttt{.detach()}}$
    \end{algorithm}
    }
\end{tcolorbox}
\begin{tcolorbox} 
[colback=white!100, colframe=SkyBlue!60, width=\textwidth, 
righttitle=0.5cm, subtitle style={boxrule=0.5pt, 
colback=yellow!50!red!25!white}, title=\textcolor{black}{\textbf{\normalsize{Algorithm 2: }}{MambaLRP in Mamba block}}]
    {
    \begin{algorithm}[H]
    \KwData{\textbf{Input:} $x$ \textcolor{ForestGreen}{(B, L, D)}}
    \KwData{\textbf{Output:} $y$ \textcolor{ForestGreen}{(B, L, D)}} 
    
        $x^\prime$: \textcolor{ForestGreen}{(B, L, E)} $\leftarrow$  \text{SiLU}(\text{Conv1d}($x$))\\
        {$g$}: \textcolor{ForestGreen}{(B, L, E)} $\leftarrow$ \text{SiLU}(\text{Linear}($x$)) \hfill\Comment{$g$ is an input-dependent gate}
        {$A$}: \textcolor{ForestGreen}{(E, N)} $\leftarrow$ Parameter \\
        {$B$}: \textcolor{ForestGreen}{(B, L, N)} $\leftarrow$  \textcolor{black}{$\text{Linear}(x^\prime)$} \\
        {$C$}: \textcolor{ForestGreen}{(B, L, N)} $\leftarrow$  \textcolor{black}{$\text{Linear}(x^\prime)$} \hfill\Comment{$C$ is input-dependent}
        {$\Delta$}: \textcolor{ForestGreen}{(B, L, E)} $\leftarrow$  $\text{Softplus}(\text{Parameter} + \text{Linear}(x^\prime))$ \\
        {$\bar{A}$}, {$\bar{B}$}: \textcolor{ForestGreen}{(B, L, E, N)} $\leftarrow$ discretize({$\Delta$}, {$A$}, {$B$}) \hfill \Comment{{$\bar{A}$} and {$\bar{B}$} are input-dependent} 
        $y_\texttt{SSM}$: \textcolor{ForestGreen}{(B, L, E)}  $\leftarrow \text{SSM}(\bar{A}\textcolor{Plum}{\texttt{.detach()}}, \bar{B}\textcolor{Plum}{\texttt{.detach()}}, C\textcolor{Plum}{\texttt{.detach()}})(x^\prime)$ \\
        $y^\prime$: \textcolor{ForestGreen}{(B, L, E)}  $\leftarrow 0.5(y_\texttt{SSM} \odot g) + 0.5[y_\texttt{SSM} \odot g]\textcolor{Plum}{\texttt{.detach()}}$ \\
        $y$: \textcolor{ForestGreen}{(B, L, D)} $\leftarrow \text{Linear}(y^\prime)$
        
        \textbf{return} $y$
    \end{algorithm}
    }
\end{tcolorbox}
The following list represents the hyperparameters of the above-mentioned algorithms:
\begin{table}[h]
    \centering
    \begin{tabular}{ll}
         B & batch size \\
         L & sequence length \\
         D & hidden dimension \\
         E & expanded hidden dimension \\
         N & SSM dimension \\
    \end{tabular}
\end{table}

Explanations generated by propagation-based methods rely on gradient computations, which can result in noisy explanations in models with many layers. This is due to the phenomena of gradient shattering and the presence of noisy gradients, which are more common in deep complex models \cite{DOMBROWSKI2022108194, achtibat2024attnlrp}. To mitigate this, we apply the generalized LRP-$\gamma$-rule to the convolution layers of the Vision Mamba model to improve the signal to noise ratio, thereby enhancing explanations. The generalized LRP-$\gamma$ rule is defined in Eq.~\ref{eq:gamma_rule}:

\begin{align}
    \label{eq:gamma_rule}
    \mathcal{R}(x_i) = 
    \begin{cases}
    \sum_j \frac{x_i^+(w_{ij} + \gamma w_{ij}^+) + x_i^-(w_{ij} + \gamma w_{ij}^-)} {\sum_{i} {x_i^+(w_{ij} + \gamma w_{ij}^+) + x_i^-(w_{ij} + \gamma w_{ij}^-)}} \mathcal{R}(y_j) & \text{if} \quad z_j > 0 \\ \\
    \sum_j \frac{x_i^+(w_{ij} + \gamma w_{ij}^-) + x_i^-(w_{ij} + \gamma w_{ij}^+)} {\sum_{i} {x_i^+(w_{ij} + \gamma w_{ij}^-) + x_i^-(w_{ij} + \gamma w_{ij}^+)}} \mathcal{R}(y_j) & \text{else}
    \end{cases}
\end{align}

where ${(.)}^+ = \text{max}(0, .)$ and ${(.)}^- = \text{min}(0, .)$, and $z_j = \sum_i x_i w_{ij}$. In our experiments, the parameter $\gamma$ is set to 0.25. Our observations reveal that applying this rule to the language models does not lead to any discernible improvements. Therefore, we use the LRP-0 rule in these models.

\subsubsection{LRP composites for Vision Mamba}
\label{sec:composites}

As mentioned in Section~\ref{sec:mambalrp_details}, we apply the generalized LRP-$\gamma$ rule to the convolution layers of the Vim-S model to produce more faithful explanations. In this experiment, we justify this choice. Vision Mamba is composed of a number of  blocks and in each block, there are several linear and convolution layers, where the generalized LRP-$\gamma$ rule can be used. As can be seen in Table \ref{tab:vim_composite_search}, the LRP-0 rule is sufficient to produce meaningful explanations. However, we can perform a hyperparameter search by applying the LRP-$\gamma$ rule across different layers of the model to find the most accurate LRP composite.

\begin{table}[!ht]
    \captionsetup{skip=0.5\baselineskip}
    \centering
    \caption{Finding the best LRP composite for Vision Mamba. The layers in which the generalized LRP-$\gamma$ rule is applied are represented with LRP-$\gamma$ and the ones in which the basic LRP rule, \ie LRP-0, is used are represented with LRP-0.}
    \small
    \begin{tabular}{ccc|c}
         \toprule
         \textbf{in-proj} & \textbf{out-proj}  & \textbf{conv1d} & {\parbox{2cm}{\centering \textbf{ImageNet} \\ (\textbf{$\Delta{A}^F \uparrow$})}}\\
         \midrule
         LRP-$\gamma$ & LRP-$\gamma$ & LRP-$\gamma$ & 4.196\\
         LRP-$\gamma$ & LRP-$\gamma$ & LRP-0 & 4.218\\
         LRP-0 & LRP-$\gamma$ & LRP-$\gamma$ & 4.283\\
         LRP-0 & LRP-$\gamma$ & LRP-0 & 4.336\\
         LRP-$\gamma$ & LRP-0 & LRP-$\gamma$ & 4.597\\
         LRP-$\gamma$ & LRP-0 & LRP-0 & 4.599\\
         LRP-0 & LRP-0 & LRP-0 & 4.684\\
         \rowcolor{cyan!10} LRP-0 & LRP-0 & LRP-$\gamma$ & \textbf{4.715}\\
         \bottomrule  
    \end{tabular}
    \label{tab:vim_composite_search}
\end{table}

We apply the LRP-$\gamma$ rule across different combinations of the input projection (in-proj), output projection (out-proj), and convolution layers of each block. Subsequently, we perform the perturbation experiment to analyze the faithfulness of each combination. We can observe that the best result can be achieved when the LRP-$\gamma$ rule is only used in convolution layers. In all of these combinations, the value of $\gamma$ is set to 0.25.

\subsection{Further details of other explanation methods}
Some of the explanation methods that we used in this study have a set of hyperparameters. Table~\ref{tab:baseline_params} provides further details on the specific values assigned to these hyperparameters, chosen based on the values suggested in the original papers \cite{intgrad, smoothgrad}.

\begin{table}[!ht]
    \captionsetup{skip=0.5\baselineskip}
    \centering
    \caption{Hyperparameters of other explanation methods. The parameters $\mu$ and $\sigma$ represent the mean and standard deviation of noise, respectively, while the parameter $m$ denotes the sample size.}
    \label{tab:baseline_params}
    \small
    \begin{tabular}{c|c}
         \toprule
         \textbf{Method} &  \textbf{Hyperparameters}\\ [0.5ex]
         \midrule
         SmoothGrad & $\mu=0$, $\sigma=0.15$, $m=30$ \\[0.5ex]
         Integrated Gradients & $m=30$ \\
         \bottomrule
    \end{tabular}
\end{table}

In the vision experiments, we used the original implementations \footnote{\url{https://github.com/AmeenAli/HiddenMambaAttn/}} of the AttnRoll and MambaAttr methods, provided to explain the Vim-S model. Given the unavailability of code for adapting these approaches to the language models, namely Mamba-130M, Mamba-1.4B, and Mamba-2.8B, we have developed our own implementation. In the vision case, the authors obtain the final relevance map by extracting the row associated with the CLS token in the attention matrix. However, since our language models lack a CLS token, we get the final relevance map from the row associated with the last token in the attention matrix. This is because predictions are based on the last state in these models. For the gradient-based methods, we use the implementations available in the Captum library\footnote{\url{https://captum.ai/}}.

\subsection{Further details on evaluation metrics}
As mentioned in Section~\ref{sec:quantitative_eval}, the \textit{flipping} and \textit{insertion} metrics can be used to evaluate the quality of the generated explanations. It is important to note that starting with unperturbed images and gradually applying perturbations until fully perturbed images are obtained yields identical results for both the \textit{flipping} and \textit{insertion} metrics. Therefore, we have only reported the results for the \textit{flipping} experiment.

In our \textit{flipping} evaluations, we calculate the area under the curve (AUC) by starting from full images and progressively masking the pixels with zeros until we reach completely masked images. The perturbation steps are defined using \code{np.linspace(0, 1, 11)}. In our vision experiments, Images are normalized using the ImageNet mean and standard deviation, and the explanations are generated and evaluated for the predicted class. Unlike \cite{ali2024hidden}, which tracks changes in the model's top-1 accuracy, we monitor changes in the output logit of the predicted class. 

\subsection{Further ablation experiments}
\paragraph{Comparing strategies for managing the Mamba block's multiplicative gate:}
\label{sec:ablation_multiplicative_gate}
In Section \ref{section:product}, we proposed several strategies to mitigate conservation violation in the Mamba block's multiplicative gate. In this experiment, we evaluate the proposed approaches. As can be seen in Table~\ref{tab:multiplicative_gate_handling}, detaching the multiplicative gate $z_B$ leads to lower faithfulness scores compared to the half-relevance propagation approach. To retain conservation, an alternative approach is to detach the SSM's output $z_A$, which limits capturing long-range dependencies, a task for which this branch is designed for. Detaching it may result in a loss of valuable information used by the model to make predictions. Therefore, this approach is not considered in Table~\ref{tab:multiplicative_gate_handling}.

\begin{table}[t]
    \centering
    \captionsetup{skip=0.5\baselineskip}
    \caption{Comparing the proposed strategies for managing the Mamba block's multiplicative gate.}
    \resizebox{0.5\textwidth}{!}{
    \begin{tabular}{l|cc}
         \toprule
         \textbf{Strategies} & \textbf{SST-2} & \textbf{ImageNet} \\
         \toprule
         Detaching the multiplicative gate $z_B$ & 1.577 & 3.592\\
         \rowcolor{cyan!10} Half-relevance propagation & \textbf{1.978} & \textbf{4.715}\\
         \bottomrule  
    \end{tabular}
    }
    \label{tab:multiplicative_gate_handling}
\end{table} 

\subsection{Additional qualitative results}
\label{sec:additional_qualitative}
In Section~\ref{sec:qualitative_results}, we qualitatively evaluated the explanations produced by MambaLRP and other baseline methods. In the following, we demonstrate further qualitative results.

\subsubsection{Natural language processing}
In the following figures, we represent explanations produced by MambaLRP and other baseline methods to interpret the Mamba-130M models trained on various datasets. In the visualizations, shades of red represent words that positively influence the model's prediction. Conversely,  shades of blue reflect negative contributions. The heatmaps of the AttnRoll and MambaAttr methods are constrained to non-negative values.
\begin{figure}[!ht]
    \centering
    \includegraphics[width=\textwidth]{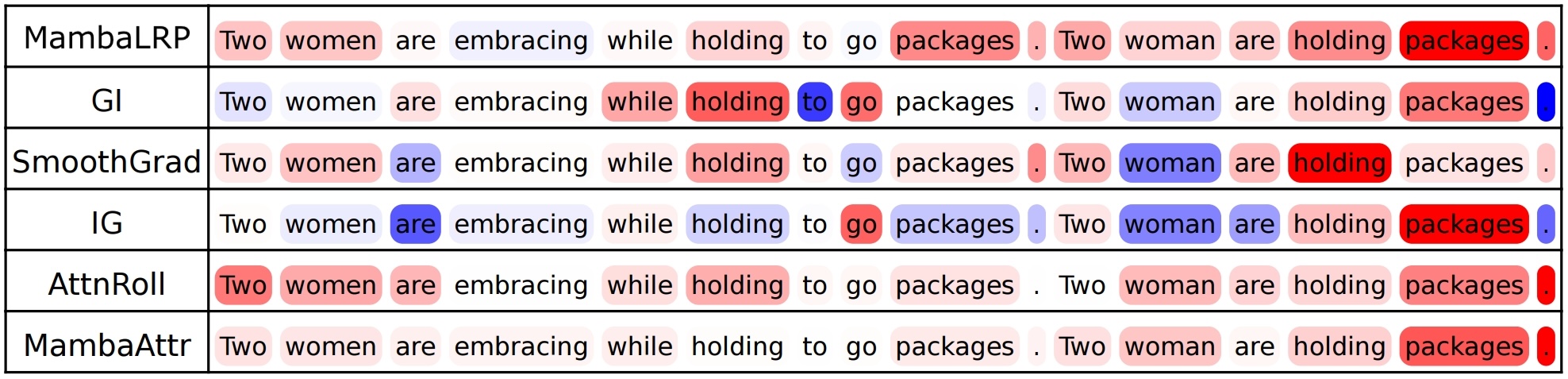}
    \caption{Explanations generated by different explanation methods for a sentence of the SNLI validation set. This sentence belongs to the `entailment' class.}
    \label{fig:snli_qualitative}
    \vspace{-8pt}
\end{figure}

\begin{figure}[!ht]
    \centering
    \includegraphics[width=\textwidth]{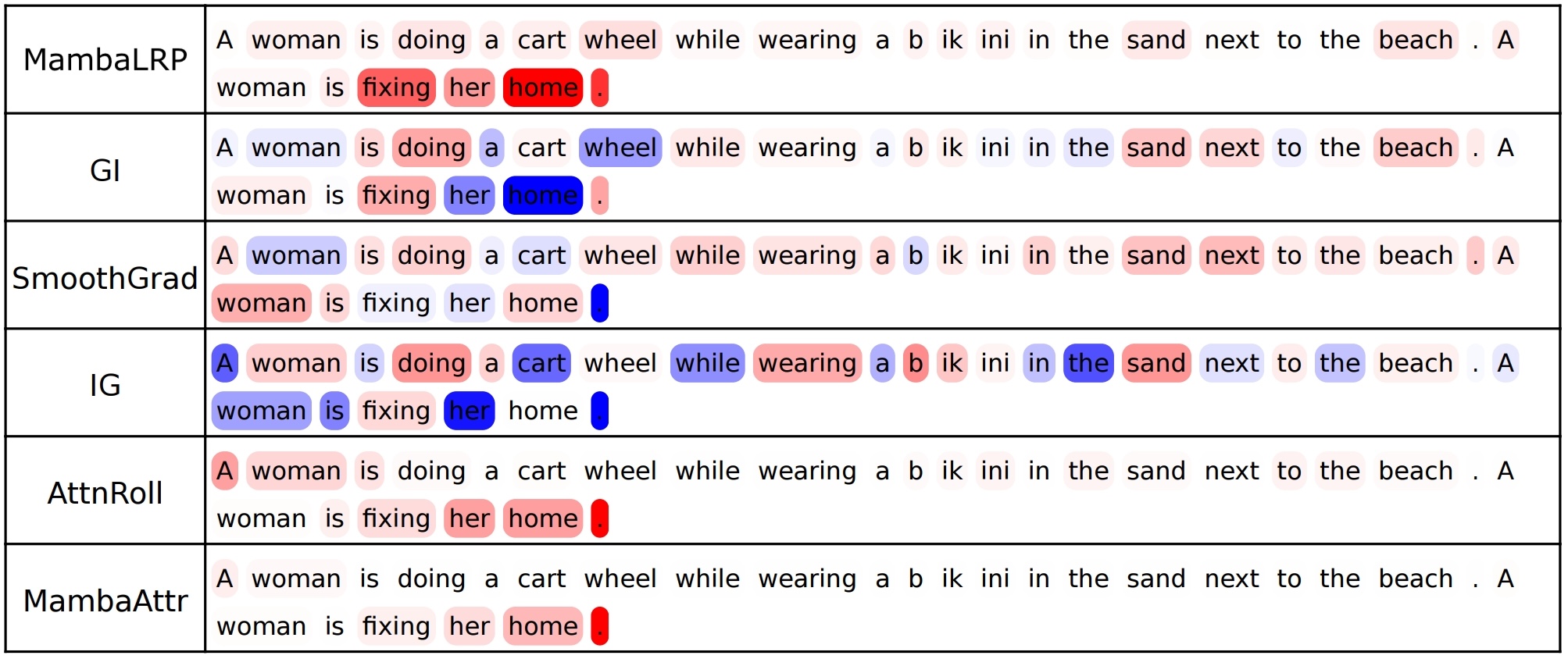}
    \caption{Explanations generated by different explanation methods for a sentence of the SNLI validation set. This sentence belongs to the `contradiction' class.}
    \label{fig:snli_qualitative_2}
    \vspace{-8pt}
\end{figure}

\begin{figure}[!ht]
    \centering
    \includegraphics[width=0.75\textwidth]{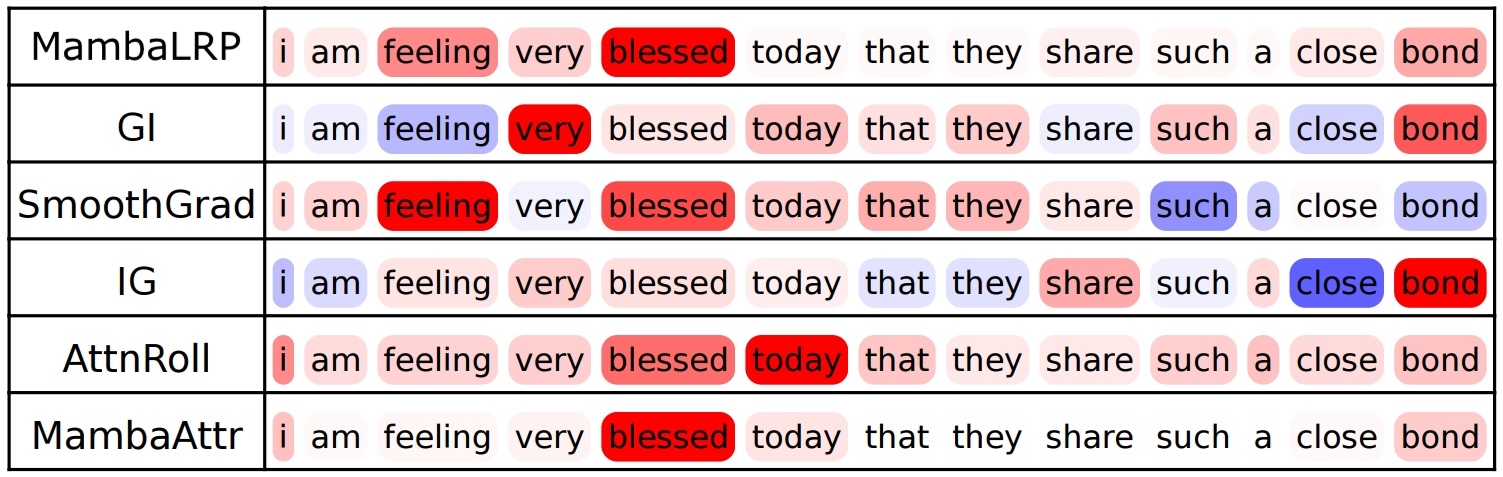}
    \caption{Explanations generated by different explanation methods for a sentence of the Emotion validation set. This sentence belongs to the `joy' class.}
    \label{fig:emotion_qualitative}
    \vspace{-8pt}
\end{figure}

\begin{figure}[!ht]
    \centering
    \includegraphics[width=\textwidth]{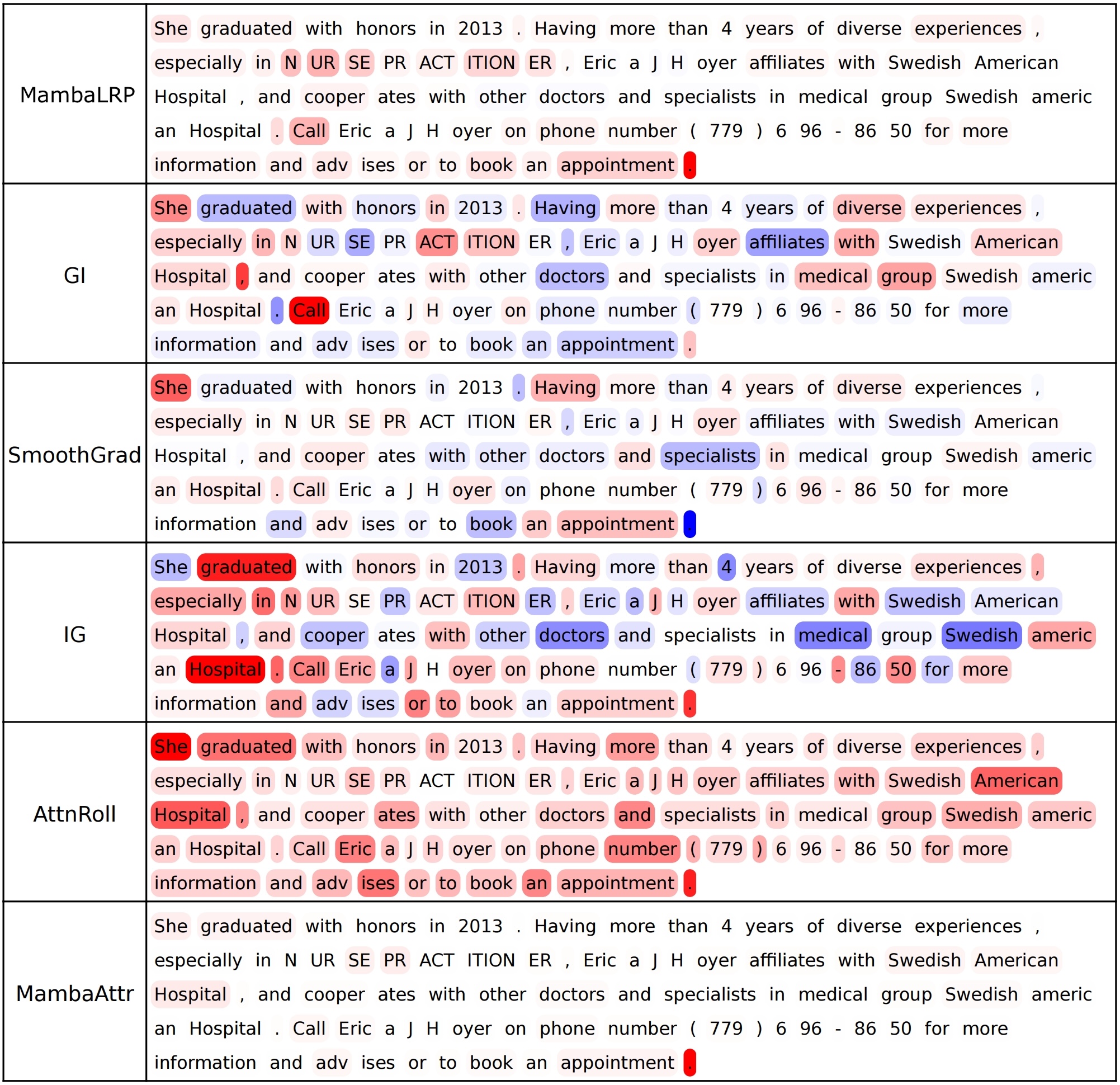}
    \caption{Explanations generated by different explanation methods for a sentence of the Medical BIOS validation set. This sentence belongs to the `nurse' class.}
    \label{fig:bios_qualitative}
    \vspace{-8pt}
\end{figure}
\subsubsection{Computer vision}
In this section, we show explanations generated by MambaLRP alongside other baseline methods to interpret the predictions of the Vim-S model on several images of the ImageNet dataset. As can be seen, explanations generated by purely gradient-based explanation methods are very noisy. In contrast, attention-based attribution methods have offered more focused and less noisy heatmaps. However, in the last two images labeled `paint brush' and `flag pole', they could not faithfully explain the model's predictions. Among these approaches, MambaLRP stands out with its ability to generate sparse explanations, offering more faithful explanations of how different image patches contribute to the final predictions.

\begin{figure}[t]
    \centering
    \includegraphics[width=\textwidth]{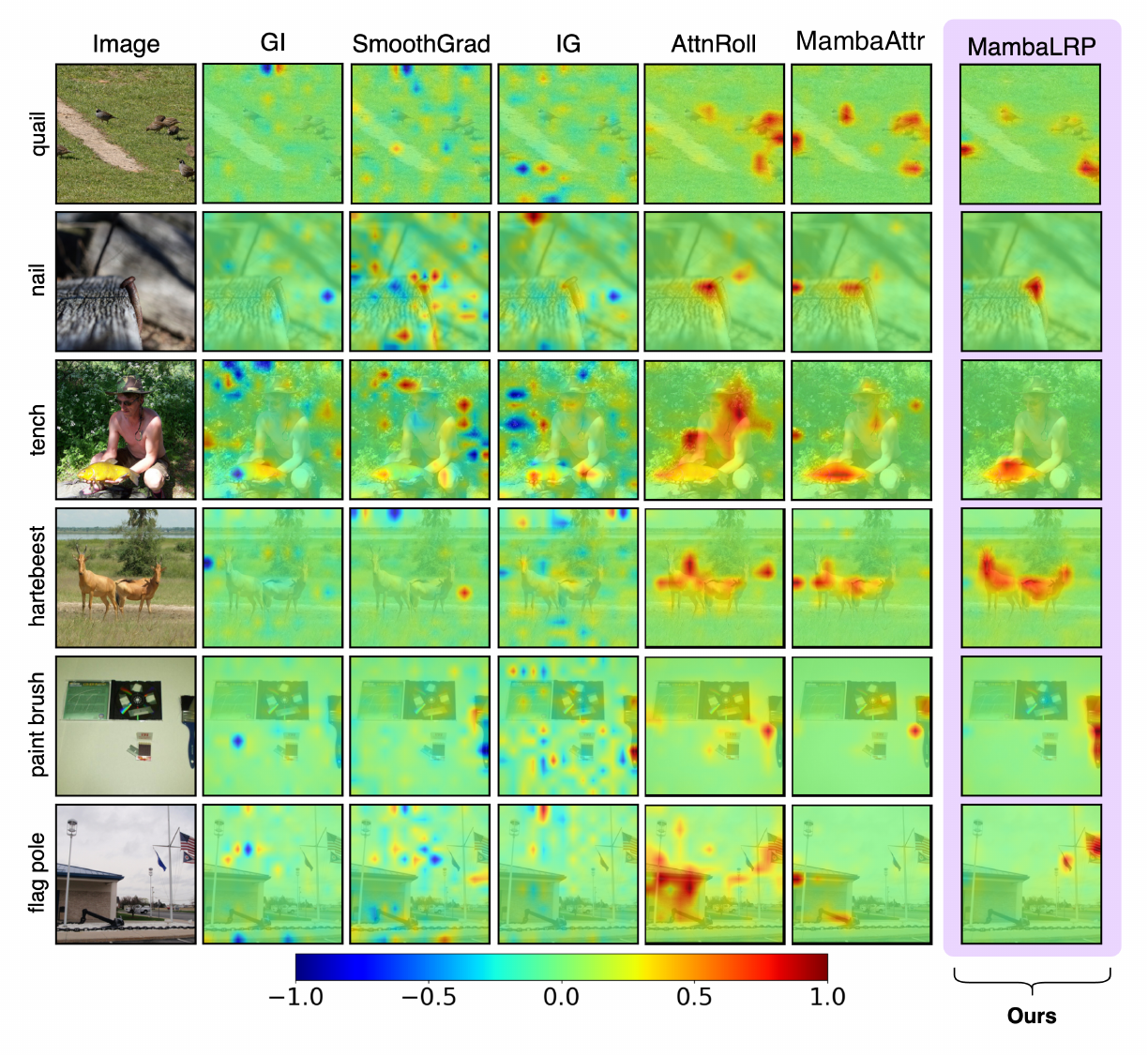}
    \caption{Explanations produced by different explanation methods for images of the ImageNet dataset. Explanations produced by AttnRoll and MambaAttr are limited to non-negative values, whereas those generated by gradient-based techniques and MambaLRP includes both positive and negative contributions.}
    \label{fig:enter-label}
\end{figure}

\subsection{Additional use case results}
\label{sec:additional_use_case}
For the needle-in-a-haystack experiment in Section \ref{sec:needle}, we use a synthetic dataset \footnote{\url{https://huggingface.co/datasets/lvwerra/needle-llama3-16x512}}. In this dataset, a single passkey (the `needle') is inserted at different locations within a collection of repeated noise sentences (the `haystack'), as described in \cite{hsieh2024ruler}. The dataset is composed of sequences with different context lengths. In our experiment, we use sequences with context lengths of 512, 1024, and 2048. We restrict the maximum context length to 2048 tokens to align with the model's training configuration, as this experiment is not designed to evaluate extrapolation beyond this limit. The goal is to focus on certain limitations of the retrieval accuracy metric and and the solutions provided by MambaLRP. We use a Mamba-2.8B model \footnote{\url{https://huggingface.co/clibrain/mamba-2.8b-chat-no_robots}}, which is finetuned on the No Robots dataset \cite{no_robots} using a context length of 2048. 
Then, we prompt the model to extract the passkey hidden among irrelevant text by completing the phrase ``The passkey is \rule{1cm}{0.15mm}". 

Retrieval accuracy is a metric, which is commonly used in the needle-in-a-haystack experiment to analyze the model's performance. The synthetic dataset used for this experiment can be designed to include misleading information, which may cause the model to generate the correct passkey based on incorrect evidence. In such cases, simply evaluating the retrieval accuracy may be insufficient. This issue can also arise when dealing with more realistic haystacks. Therefore, we introduced explanation-based retrieval accuracy (XRA) in Section~\ref{sec:use_cases}. MambaLRP and the XRA metric designed upon it can help to better examine the evidence the model relies on to retrieve the needle. In this approach, we first identify the positions of the top-K relevant tokens by MambaLRP, and then, calculate the accuracy by comparing those positions to the needle's position. We set the value of K to 2. This is because MambaLRP identifies the token immediately preceding the generated token as the most important one in most of the examples and the evidence used for the passkey retrieval is usually the second most important token.

\begin{figure}[!ht]
    \centering
    \includegraphics[width=\textwidth]{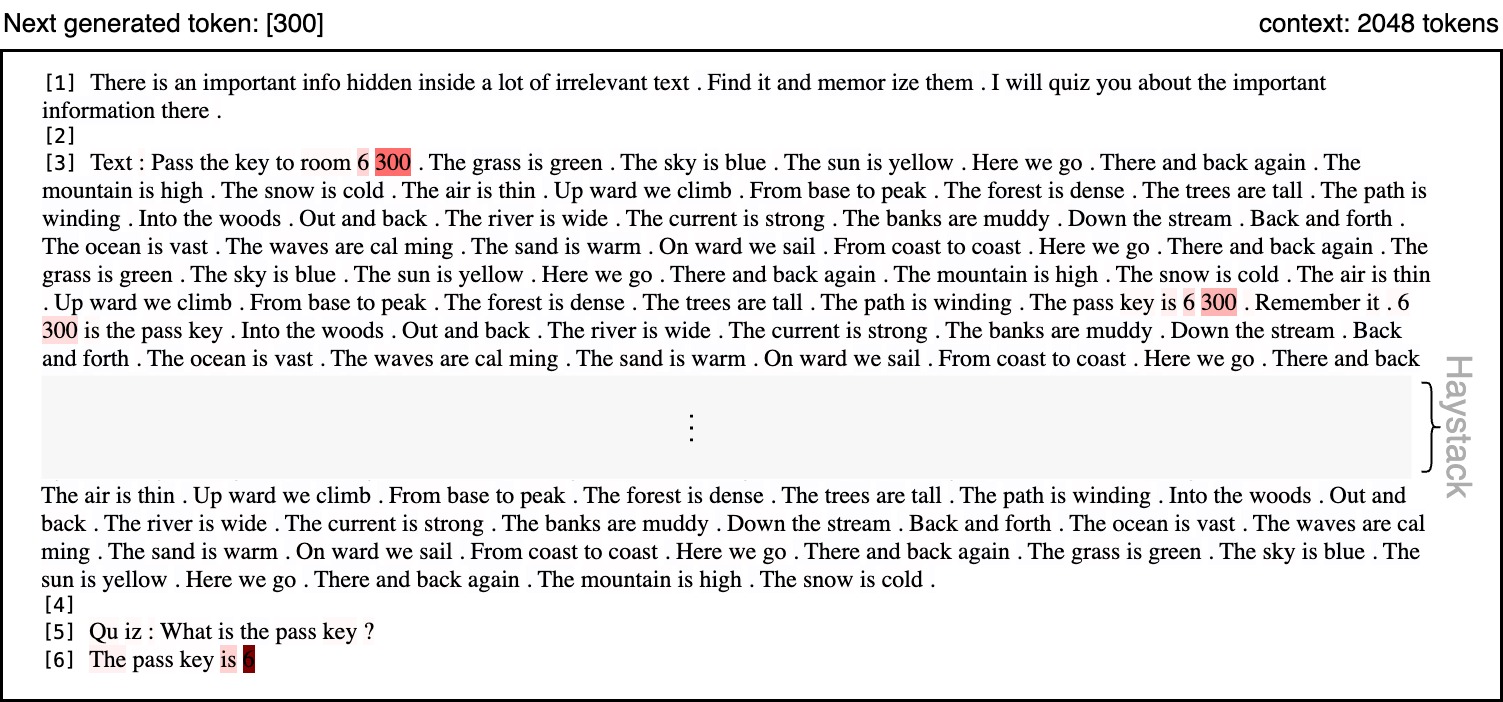}
    \caption{Detecting Clever-Hans effect in the needle-in-a-haystack test. Given the 2K context length in this example, visualizing the entire text could be confusing. Therefore, we have removed most of the haystack from the visualization. In this example, the model has generated the correct passkey but the generation is not based on truly relevant information in the text.}
    \label{fig:haystack_additional}
\end{figure}

The sample in Fig.~\ref{fig:haystack_additional} represents a scenario where our XRA approach proves valuable. In this case, the next token generated by the model is the second part of the correct passkey (300). However, the model has incorrectly focused on the number 300 in the phrase ``Pass the key to room 6300" to generate this token. Simply looking at the retrieved token might suggest that the model successfully retrieved the correct information. However, examining the MambaLRP's explanation heatmaps provides deeper insights into the model's behavior. This helps us to debug the model more effectively and design better tests to analyze its capabilities.

\subsection{Long-range dependeny comparison to Transformers}
\label{sec:lrd_transformers}

To explore the capabilities of different model architectures in handling long-range dependencies, we performed a direct comparison between Mamba and state-of-the-art Transformers (Llama-2\footnote{\url{https://huggingface.co/meta-llama/Llama-2-7b-hf}} and Llama-3\footnote{\url{https://huggingface.co/meta-llama/Meta-Llama-3-8B}}), focusing on their performance with inputs exceeding typical context lengths. 

For Llama-2 and Llama-3, we extract attributions using LRP for Transformers \cite{DBLP:conf/icml/AliSEMMW22}. As in the Mamba experiment, we generate 10 additional tokens from the HotPotQA dataset input and explain the prediction for each generated token. The results are shown in Table~\ref{fig:app_lrd}.
For Llama-2, which was trained with a context length of 4096, the generated text becomes increasingly less sensible and repetitive for contexts longer than 4k, a limitation noted also in \cite{chen2024clex, pmlr-v235-an24b}. When analyzing the histogram distribution over tokens considered relevant to predict the next token, it appears that Llama-2 uses information more uniformly across the entire context and identifies more relevant long-range dependent tokens compared to Llama-3 and Mamba. However, as presented in Table~\ref{fig:app_lrd},  its output becomes nonsensical for context lengths exceeding 4K tokens, characterized by the use of rare vocabulary and repeated tokens. Thus, the identified relevant tokens are mostly non-semantic such as new line token `$<$0x0A$>$' in Llama-2 and beginning of sentence token `$<$s$>$' found at the start of the context paragraphs. For Llama-3 and Mamba, the attributions can identify meaningful relevant tokens. When directly compared, Llama-3 uses information from more intermediate mid-range dependencies than Mamba, though both favor tokens close to the end of the input as relevant. Given Llama-3's much larger size (8B) compared to Mamba (130M) and their different training settings, this analysis supports that Mamba indeed uses long-range information. We also find that this ability is not exclusive to SSMs and can in principle also be achieved by Transformer models. 

To what extent these findings depend on the amount of training data and model complexity remains an open research question. Our investigation of long-range dependencies in recent sequence generation models highlights the value of faithful attribution methods like MambaLRP in examining the capabilities and mechanisms utilized by models during generation.

\begin{table}[!ht]
\scriptsize
\begingroup
\fontencoding{T2A}\selectfont 
\begin{tabularx}{\textwidth}{P{0.02\linewidth} P{0.04\linewidth} C C C}
\toprule
id & length & Llama-2 & Llama-3 & Mamba \\
\midrule
3 & 1k & Summary <0x0A> The ▁genus ▁D ict y os per ma ▁is ▁a & Summary Ċ The genus Dict y os per ma is a mon & Summary Ċ The species is a member of the genus Ap oll \\
\midrule
41 & 4k & Summary <0x0A> The ▁Ohio , ▁Ohio , ▁Ohio , ▁Ohio , ▁Ohio & Summary Ċ The following is a summary of Finn 's head coaching & Summary Ċ The following is a summary of the history of the \\
\midrule
109 & 8k & Summary <0x0A> The Љ Љ Љ Љ Љ Љ Љ Љ Љ & Summary Ċ The mall is anchored by Hudson 's Bay , Walmart & Summary Ċ The mall is located in the heart of the city \\
\bottomrule
& & \begin{minipage}{0.27\textwidth}
    \centering
    \includegraphics[width=\linewidth]{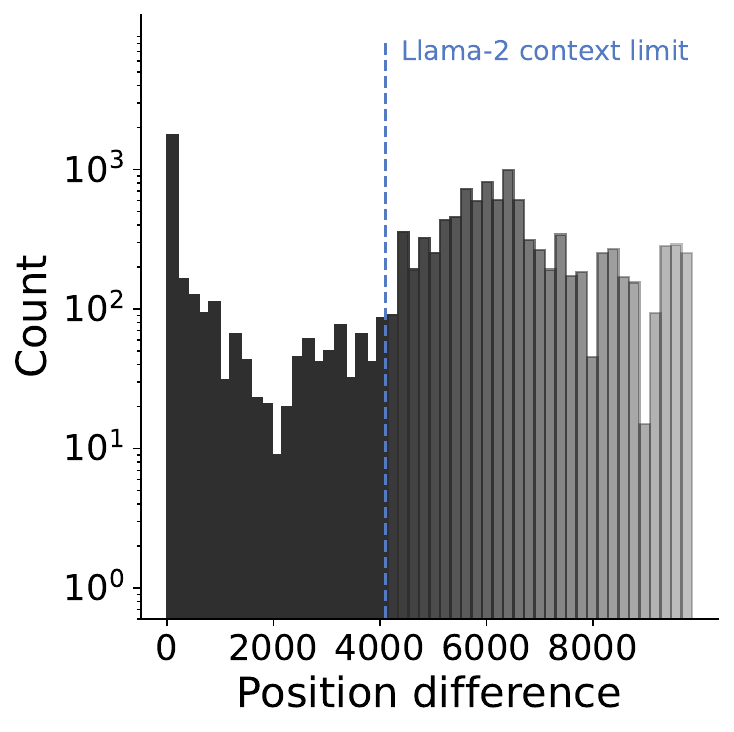}
\end{minipage} &

\begin{minipage}{0.27\textwidth}
    \centering
    \includegraphics[width=\linewidth]{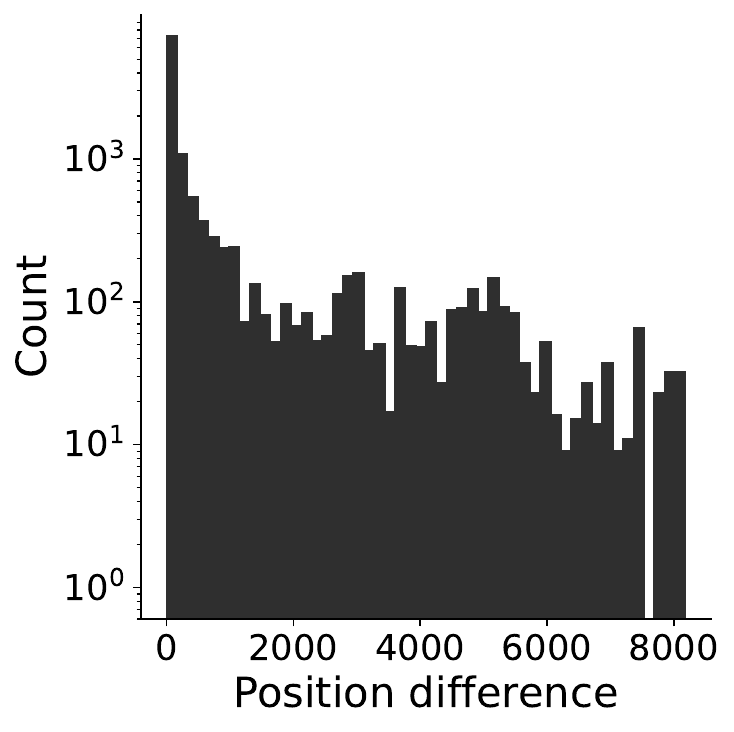}
\end{minipage} &

\begin{minipage}{0.27\textwidth}
    \centering
    \includegraphics[width=\linewidth]{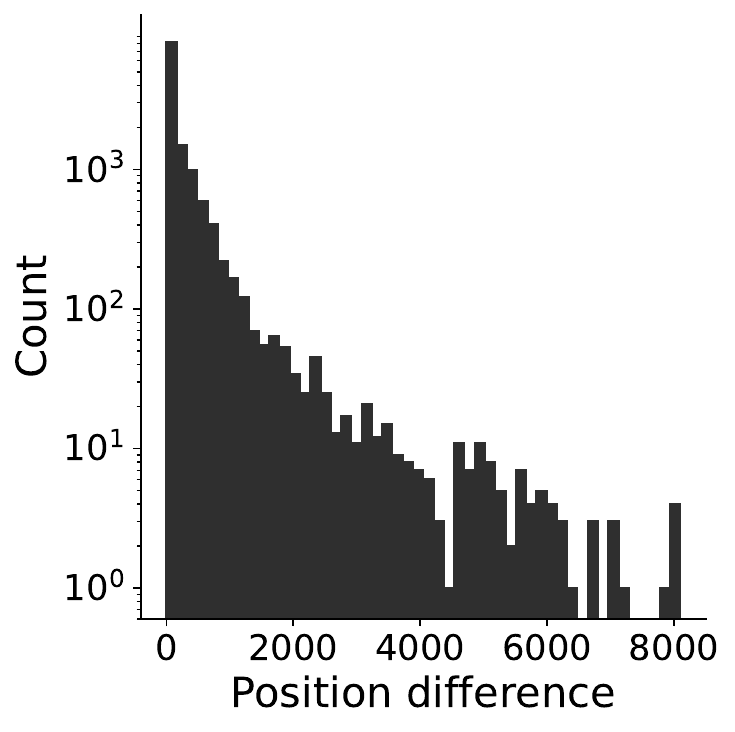}
\end{minipage} \\
\bottomrule
\end{tabularx}\\
\endgroup
\vspace{5pt}
\caption{Long-Range dependency experiment, comparing Transformers and Mamba for different context lengths.}
\label{fig:app_lrd}
\end{table}

\subsection{Runtime comparison}
\label{sec:runtime}
In this section, we report the time required for each explanation method to generate its respective explanation. These times, measured in seconds, are averaged over samples from the Medical BIOS dataset. All baseline methods are evaluated on a single A100-40GB GPU with a batch size of 1. All methods are applied to the Mamba-130M model. The results without fast CUDA kernels are shown in Table~\ref{tab:app_runtime}, while the results with fast CUDA kernels are presented in Table~\ref{tab:app_runtime_fast}. We can observe that the runtime of MambaLRP is comparable to Gradient$\times$Input. Since algorithms like Integrated Gradients and SmoothGrad require multiple function evaluations, their runtimes are significantly higher than MambaLRP and Gradient$\times$Input.

\begin{minipage}[!ht]{0.45\textwidth}
    \captionsetup{skip=0.5\baselineskip}
    \captionof{table}{Runtime comparison. The time needed for each baseline method to generate its explanations. The times, measured in seconds, are averaged over the samples from the Medical BIOS dataset. The model used in this experiment is Mamba-130M \underline{\textbf{without using}} fast CUDA kernels.}
    \label{tab:app_runtime}
    \centering
    \small
    \begin{tabular}{ l | c }
    \toprule
    \textbf{Methods} & \textbf{Runtime} \\ \midrule
        Gradient $\times$ Input & 0.7556 \\
        SmoothGrad & 22.9772\\
        Integrated Gradients & 22.8071\\
        \midrule
        AttnRoll & 2.1558 \\
        MambaAttr & 2.6661 \\
        \midrule
        MambaLRP & 0.4345 \\
    \bottomrule
    \end{tabular}
\end{minipage}
\hfill
\begin{minipage}[!ht]{0.45\textwidth}
    \captionsetup{skip=0.5\baselineskip}
    \captionof{table}{Runtime comparison. The time needed for each baseline method to generate its explanations. The times, measured in seconds, are averaged over the samples from the Medical BIOS dataset. The model used in this experiment is Mamba-130M \underline{\textbf{using}} fast CUDA kernels.}
    \label{tab:app_runtime_fast}
    \centering
    \small
    \begin{tabular}{ l | c }
    \toprule
    \textbf{Methods} & \textbf{Runtime} \\ \midrule
        Gradient $\times$ Input & 0.0335 \\
        SmoothGrad & 0.9785 \\
        Integrated Gradients & 0.9742\\
        \midrule
        AttnRoll & - \\
        MambaAttr & - \\
        \midrule
        MambaLRP & 0.0306 \\
    \bottomrule
    \end{tabular}
\end{minipage}
\end{document}